\documentclass[journal]{IEEEtran}
\usepackage{amsmath,amsfonts}
\usepackage{array}
\usepackage{algorithmic}
\usepackage[ruled,lined,linesnumbered,ruled]{algorithm2e}  
\usepackage{multirow}
\usepackage{graphicx} 
\usepackage{float} 
\usepackage[caption=false,font=footnotesize,labelfont=rm,textfont=rm]{subfig}
\usepackage{CJKutf8}
\usepackage{textcomp}
\usepackage{stfloats}
\usepackage{url}
\usepackage{verbatim}
\usepackage{graphicx}
\usepackage{cite}
\usepackage{lipsum}
\begin{document}
\begin{CJK}{UTF8}{gbsn}
\title{\textbf{Deep Recurrent Stochastic Configuration Networks for Modelling Nonlinear Dynamic Systems}}

\author{
  Gang Dang \\
  State Key Laboratory of Synthetical Automation for Process Industries \\
  Northeastern University, Shenyang 110819, China\\
   Dianhui Wang 
  \thanks{\textit{\underline{Corresponding author}}: 
\textbf{dh.wang@deepscn.com}}\\
  State Key Laboratory of Synthetical Automation for Process Industries \\
  Northeastern University, Shenyang 110819, China \\
 Research Center for Stochastic Configuration Machines\\
  China University of Mining and Technology, Xuzhou 221116, China\\
}
\maketitle
\newtheorem{remark}{\bf Remark}      

\begin{abstract}                          
 Deep learning techniques have shown promise in many domain applications. This paper proposes a novel deep reservoir computing framework, termed deep recurrent stochastic configuration network (DeepRSCN) for modelling nonlinear dynamic systems. DeepRSCNs are incrementally constructed, with all reservoir nodes directly linked to the final output. The random parameters are assigned in the light of a supervisory mechanism, ensuring the universal approximation property of the built model. The output weights are updated online using the projection algorithm to handle the unknown dynamics. Given a set of training samples, DeepRSCNs can quickly generate learning representations, which consist of random basis functions with cascaded input and readout weights. Experimental results over a time series prediction, a nonlinear system identification problem, and two industrial data predictive analyses demonstrate that the proposed DeepRSCN outperforms the single-layer network in terms of modelling efficiency, learning capability, and generalization performance.
\end{abstract}


\section{Introduction}
\IEEEPARstart{N}{owdays}, deep neural networks (DNNs) have gained considerable attention in complex modelling tasks due to their effectiveness and efficiency in data representation \cite{ref1,ref2,ref201}. However, for real-world industrial applications, fluctuations in the controlled plant and external disturbances can lead to unknown dynamic orders, making it challenging to build accurate DNN models \cite{ref202,ref203}. Recurrent neural networks (RNNs) have feedback connections between neurons, which can handle the uncertainty caused by the selected input variables that are fed into a learner model. To further capture the temporal structure of dynamic process data, researchers have integrated deep learning techniques into the RNN and established a hierarchical learning framework, termed DeepRNN for better modelling performance \cite{ref3,ref4}. Unfortunately, DeepRNNs utilize the error back-propagation (BP) algorithm to train the network connection weights and biases, which suffer from the sensitivity to learning rate, slow convergence, and local minima. Reservoir computing (RC) provides an alternative approach to train RNNs \cite{ref6,ref7,ref711,ref712}. As a class of randomized learning algorithms, RC introduces a fixed, sparsely connected reservoir to capture the temporal dynamics. Only the output weights are calculated using the least squares method, effectively addressing the limitations associated with the BP-based methods. Nowadays, it has been widely applied to train various randomized learner models, such as echo state networks (ESNs) \cite{ref8} and liquid state machines (LSMs) \cite{ref9}. This paper focuses on the development of the deep RC-based framework.

Early in \cite{ref001}, a formal analysis was conducted to assess the representation power and learning complexity of neural networks, leading to a common sense within the community that DNNs possess greater expressiveness compared to shallow networks. Deep echo state networks (DeepESNs) can be regarded as an extension of ESNs by incorporating multi-layer reservoirs, which not only allow for hierarchical feature representation across different levels but also maintain the efficient training process characteristic of RC algorithms \cite{ref10,ref11}. In \cite{ref12}, a multi-step learning ESN was proposed for nonlinear time series prediction, verifying that DeepESNs require less computational cost than the standard ESNs to achieve superior or comparable performance. Furthermore, Gallicchio et. al extended the unique echo state property (ESP) of ESNs to deep frameworks in \cite{ref13} and summarized the advancements in DeepESN development, analysis, and application in \cite{ref14}. However, these aforementioned models lack a basic understanding on both the structure setting and the random parameters assignment. According to \cite{ref16,ref17}, a deep network exhibits excellent learning and generalization performance when it is incrementally constructed using a data-dependent random parameter scope and its convergence satisfies certain conditions. 

In 2017, Wang and Li pioneered an advanced randomized learner model, termed stochastic configuration networks (SCNs) \cite{ref18}, which assign the random weights and biases in the light of a supervisory mechanism. Built on the SCN concept, in \cite{ref181}, we introduced a recurrent version of SCNs (RSCNs) to create a class of randomized universal approximators for temporal data. This paper further develops the RSCN model and establishes the universal approximation property for its deep version, termed DeepRSCN. DeepRSCN starts with a small network size and stochastically configures the current reservoir layer based on the recurrent stochastic configuration (RSC) algorithm. Then, this process is extended to deeper layers until meeting the terminal conditions. Finally, an online update of the output weights by using the projection algorithm is performed to adapt to dynamic changes in the systems. The main contributions of this work can be summarized as follows.

\begin{itemize}
  \item [1)] 
 A novel deep learning framework with multiple reservoir layers is proposed, effectively enhancing the model's feature representation capabilities and improving the learning speed.      
  \item [2)]
Each node is directly linked to the final output and generated from an adjustable scale interval in the light of the supervisory mechanism, solving the problems of structure design and random parameters selection, and theoretically guaranteeing the universal approximation property of the built model.
  \item [3)]
The developed framework is applied to various nonlinear dynamic modelling tasks. A comprehensive comparison with the original RSCNs, ESNs, and their deep variants demonstrates the superiority of the proposed DeepRSCNs in terms of computational efficiency, learning capability, and generalization performance.
\end{itemize}

The remainder of this paper is organized as follows. Section 2 reviews the related knowledge about SCNs, ESNs, and DeepESNs. Section 3 details the proposed DeepRSCNs and provides the theoretical analysis. Section 4 presents the experimental results and comparisons. Finally, Section 5 concludes this paper.

\section{Technical support}
\subsection{Stochastic configuration networks}
SCNs are a class of randomized learner models with universal approximation properties \cite{ref18}, which exploit the stochastic configuration algorithm to train the network, effectively avoiding the complex iterative adjustment. The SCN construction can be easily implemented, and address the the problems of network parameter selection and arbitrary structure determination in other randomized neural networks. Due to their merits of high learning efficiency, less human intervention, and strong approximation ability, SCNs have shown great potential in modelling nonlinear dynamic systems \cite{ref17,ref18,ref19,ref20}. For more details about SCNs, one can refer to \cite{ref18}.

Given an objective function $f:{{\mathbb{R}}^{K}}\to {{\mathbb{R}}^{L}}$ and ${n}_{\max }$ groups of training samples $\left\{ \mathbf{u}\left( n \right),\mathbf{t}\left( n \right) \right\}$, $\mathbf{u}\left( n \right)=\left[ {{u}_{1}}\left( n \right),\ldots ,{{u}_{K}}\left( n \right) \right]^{\top}$, $\mathbf{t}\left( n \right)=\left[ {{t}_{1}}\left( n \right),\ldots ,{{t}_{L}}\left( n \right) \right]^{\top}$, let $\mathbf{U}=\left[ \mathbf{u}\left( 1 \right),\ldots ,\mathbf{u}\left( {{n}_{\max }} \right) \right]\in {{\mathbb{R}}^{K\times {{n}_{\max }}}}$ and ${\bf{T}} = \left[ {{\bf{t}}\left( 1 \right), \ldots ,} \right.$ $\left. {{\bf{t}}\left( {{n_{\max }}} \right)} \right] \in {^{L \times {n_{\max }}}}$ denote the input data and the corresponding output, respectively. Suppose we have built a single-layer feedforward network with $N-1$ hidden nodes, that is, 
\begin{equation}
\label{eq1}
{{f}_{N-1}}=\sum\limits_{j=1}^{N-1}{{{\mathbf{\beta }}_{j}}{{g}_{j}}\left( \mathbf{w}_{j}^{\top }\mathbf{U}+{{\mathbf{b}}_{j}} \right)}\left( {{f}_{0}}=0,N=1,2,3... \right),
\end{equation}
where ${{\mathbf{w}}_{j}}$, ${{\mathbf{b}}_{j}}$, and ${{\mathbf{\beta }}_{j}}$ are the input weight, bias, and output weight of the $j\text{-th}$ hidden node, ${{g}}$ is the activation function. The residual error between the current model output and the expected output ${{f}_{\text{exp}}}$ is defined as
\begin{equation}
\label{eq2}
{{e}_{N-1}}={{f}_{\text{exp}}}-{{f}_{N-1}}=\left[ {{e}_{N-1,1}},{{e}_{N-1,2}},...{{e}_{N-1,L}} \right].
\end{equation}
If ${{e}_{N-1}}$ fails to meet the preset error tolerance $\varepsilon $, the network automatically configures a new random basis function ${{g}_{N}}$ in the light of the supervisory mechanism. Specifically, to prevent over-fitting, an additional condition is introduced for stopping the nodes adding and a step size ${{N}_{\text{step}}}$ (${{N}_{\text{step}}}<N$) is used in the following early stopping criterion:
\begin{equation}
\label{eq201}
{\left\| {{e_{{\rm{val}},N - {N_{{\rm{step}}}}}}} \right\|_F} \le {\left\| {{e_{{\rm{val}},N - {N_{{\rm{step}}}} + 1}}} \right\|_F} \le  \ldots  \le {\left\| {{e_{{\rm{val}},N}}} \right\|_F},
\end{equation}
where ${{e}_{\text{val},N}}$ is the validation residual error with $N$ hidden nodes and ${{\left\| \bullet  \right\|}_{F}}$ represents the $F$ norm. If Eq. (\ref{eq201}) is satisfied, the number of hidden nodes will be set to $N-{{N}_{\text{step}}}$. By evaluating the newly constructed network, the residual error can be updated until satisfying the terminal conditions. \\

\textbf{Theorem 1 \cite{ref18}.} Let span($\Gamma$) be dense in ${{L}_{2}}$ space and $\forall g\in \Gamma ,0<\left\| g \right\|<{{b}_{g}}$, ${{b}_{g}}\in {{\mathbb{R}}^{+}}$. Given $0<r<1$ and a non-negative real number sequence $\left\{ {{\mu }_{N}} \right\}$, satisfy $\underset{N\to \infty }{\mathop{\lim }}\,{{\mu }_{N}}=0$ and ${{\mu }_{N}}\le (1-r)$. For $N\text{=}1,2,...$, define
\begin{equation}
\label{eq3}
{{\delta }_{N}}=\sum\limits_{q=1}^{L}{\delta _{N,q}^{{}}},\delta _{N,q}^{{}}=(1-r-{{\mu }_{N}}){{\left\| {{e}_{N-1,q}} \right\|}^{2}},q=1,2,\ldots ,L.
\end{equation}
If the generated random basis function ${{g}_{N}}$ satisfies the following inequality constraint:
\begin{equation}
\label{eq4}
{{\left\langle {{e}_{N-1,q}},{{g}_{N}} \right\rangle }^{2}}\ge b_{g}^{2}\delta _{N,q}^{{}},q=1,2,\ldots ,L,
\end{equation}
and the output weight is determined by the global least square method, that is, 
\begin{equation}
\label{eq5}
\left[ \mathbf{\beta }_{1}^{*},\mathbf{\beta }_{2}^{*},...,\mathbf{\beta }_{N}^{*} \right]=\underset{\mathbf{\beta }}{\mathop{\arg \min }}\,\left\| {{f}_{\exp }}-\sum\limits_{j=1}^{N}{{{\mathbf{\beta }}_{j}}{{g}_{j}}} \right\|.
\end{equation}
Then, we can obtain $\underset{N\to \infty }{\mathop{\lim }}\,\left\| {{f}_{\text{exp}}}-{{f}_{N}} \right\|=0$.

\subsection{Echo state network and its deep version}
ESNs utilize the reservoir to transform input signals into a high-dimensional state space, with the readout being calculated through a linear combination of the reservoir states. The random parameters are generated from a fixed uniform distribution and only the output weights need to be trained. 

Consider an ESN model:
\begin{equation} \label{eq6}
{\bf{x}}(n) = g({{\bf{W}}_{{\rm{in}}}}{\bf{u}}(n) + {{\bf{W}}_{\mathop{\rm r}\nolimits} }{\bf{x}}(n - 1) + {\bf{b}}),
\end{equation}
\begin{equation} \label{eq7}
{\bf{y}}(n) = {{\bf{W}}_{\rm{out}}}\left( {{\bf{x}}(n),{\bf{u}}(n)} \right),
\end{equation}
where $\mathbf{u}(n)\in {{\mathbb{R}}^{K}}$ is the input signal; $\mathbf{x}(n)\in {{\mathbb{R}}^{N}}$ is the internal state of the reservoir; ${{\mathbf{W}}_{\text{in}}}\in {{\mathbb{R}}^{N\times K}},{{\mathbf{W}}_{\rm r}}\in {{\mathbb{R}}^{N\times N}}$ represent the input and reservoir weight matrices, respectively; $\mathbf{b}$ is the bias; ${{\mathbf{W}}_{\rm{out}}}\in {{\mathbb{R}}^{L\times \left( N+K \right)}}$ is the output weight matrix; $K$ and $L$ are the dimensions of input and output; and $g$ is the reservoir activation function. ${{\mathbf{W}}_{\text{in}}},{{\mathbf{W}}_{\text{r}}},\mathbf{b}$ are selected from the uniform distribution $\left[ -\lambda ,\lambda  \right].$ Notably, the value of $\lambda $ has a significant impact on model performance. Researchers have focused on optimizing the weight-scaling factor, and some promising results have been reported in \cite{ref21,ref22}. However, the optimization process inevitably increases the complexity of the algorithm. Therefore, selecting a data-dependent and adjustable $\lambda $ is essential.

Define $\mathbf{X}\text{=}\left[ \left( \mathbf{x}(1),\mathbf{u}(1) \right),\ldots ,\left( \mathbf{x}({{n}_{\max }}),\mathbf{u}({{n}_{\max }}) \right) \right]$, where ${{n}_{\max }}$ is the number of training samples. The model output is ${\bf{Y}} = \left[ {{\bf{y}}\left( 1 \right),{\bf{y}}\left( 2 \right),...{\bf{y}}\left( {{n_{max}}} \right)} \right] = {\bf{W}}_{{\mathop{\rm out}\nolimits} }^{}{\bf{X}}$. If the desired output $\mathbf{T}=\left[ \mathbf{t}\left( 1 \right),\mathbf{t}\left( 2 \right),...\mathbf{t}\left( {{n}_{max}} \right) \right]$ is known, ${{\mathbf{W}}_{\rm{out}}}$ can be calculated by the least squares method, that is, 
\begin{equation} \label{eq9}
{\bf{W}}_{{\mathop{\rm out}\nolimits} }^ \top  = {\left( {{\bf{X}}{{\bf{X}}^ \top }} \right)^{ - 1}}{\bf{X}}{{\bf{T}}^ \top }.
\end{equation}
$\mathbf{x}(0)$ is usually initialized as a zero matrix and a few warm-up samples are employed to minimize the influence of the initial zero states.

In \cite{ref12}, a DeepESN framework is introduced, where the output of the previous reservoir and the network input $\mathbf{u}\left( n \right)$ are combined as the input of the current reservoir. Let ${{\mathbf{x}}^{\left( i \right)}}(n),\mathbf{W}_{\text{in}}^{\left( i \right)},\mathbf{W}_{\text{r}}^{\left( i \right)},\mathbf{W}_{\text{out}}^{\left( i \right)},{{\mathbf{b}}^{\left( i \right)}},{{\mathbf{Y}}^{\left( i \right)}}$ denote the reservoir state, input weight, reservoir weight, output weight, bias, and output of the $i\text{-th}$ reservoir. Define ${{\mathbf{X}}^{\left( j \right)}}\text{=}\left[ \left( {{\mathbf{x}}^{\left( j \right)}}\left( 1 \right),\mathbf{u}\left( 1 \right) \right),\ldots ,\left( {{\mathbf{x}}^{\left( j \right)}}\left( {{n}_{\max }} \right),\mathbf{u}\left( {{n}_{\max }} \right) \right) \right]$, and the model output can be calculated by

\begin{footnotesize}
\begin{equation} \label{eq10}
\begin{array}{*{20}{c}}
{\left\{ {\begin{array}{*{20}{c}}
{{{\bf{x}}^{\left( 1 \right)}}(n) = g({\bf{W}}_{{\rm{in}}}^{\left( 1 \right)}{\bf{u}}(n) + {\bf{W}}_{\rm{r}}^{\left( 1 \right)}{{\bf{x}}^{\left( 1 \right)}}(n - 1) + {{\bf{b}}^{\left( 1 \right)}})}\\
\begin{array}{l}
{\bf{W}}_{{\mathop{\rm out}\nolimits} }^{\left( 1 \right) \top } = {\left( {{{\bf{X}}^{\left( 1 \right)}}{{\bf{X}}^{\left( 1 \right) \top }}} \right)^{ - 1}}{{\bf{X}}^{\left( 1 \right)}}{{\bf{T}}^ \top }\\
{{\bf{Y}}^{\left( 1 \right)}} = {\bf{W}}_{{\mathop{\rm out}\nolimits} }^{\left( 1 \right)}{{\bf{X}}^{\left( 1 \right)}}
\end{array}
\end{array},} \right.}\\
{\left\{ {\begin{array}{*{20}{c}}
{{{\bf{x}}^{\left( 2 \right)}}(n) = g({\bf{W}}_{{\rm{in}}}^{\left( 2 \right)}\left[ {\begin{array}{*{20}{c}}
{{{\bf{Y}}^{\left( 1 \right)}}^ \top }\\
{{\bf{u}}(n)}
\end{array}} \right] + {\bf{W}}_{\rm{r}}^{\left( 2 \right)}{{\bf{x}}^{\left( 2 \right)}}(n - 1) + {{\bf{b}}^{\left( 2 \right)}})}\\
\begin{array}{l}
{\bf{W}}_{{\mathop{\rm out}\nolimits} }^{\left( 2 \right) \top } = {\left( {{{\bf{X}}^{\left( 2 \right)}}{{\bf{X}}^{\left( 2 \right) \top }}} \right)^{ - 1}}{{\bf{X}}^{\left( 2 \right)}}{{\bf{T}}^ \top }\\
{{\bf{Y}}^{\left( 2 \right)}} = {\bf{W}}_{{\mathop{\rm out}\nolimits} }^{\left( 2 \right)}{{\bf{X}}^{\left( 2 \right)}}
\end{array}
\end{array},} \right.}\\
 \cdots \\
{\left\{ {\begin{array}{*{20}{c}}
{{{\bf{x}}^{\left( S \right)}}(n) = g({\bf{W}}_{{\rm{in}}}^{\left( S \right)}\left[ {\begin{array}{*{20}{c}}
{{{\bf{Y}}^{\left( {S - 1} \right)}}^ \top }\\
{{\bf{u}}(n)}
\end{array}} \right] + {\bf{W}}_{\rm{r}}^{\left( S \right)}{{\bf{x}}^{\left( S \right)}}(n - 1) + {{\bf{b}}^{\left( S \right)}})}\\
\begin{array}{l}
{\bf{W}}_{{\mathop{\rm out}\nolimits} }^{\left( S \right) \top } = {\left( {{{\bf{X}}^{\left( S \right)}}{{\bf{X}}^{\left( S \right) \top }}} \right)^{ - 1}}{{\bf{X}}^{\left( S \right)}}{{\bf{T}}^ \top }\\
{{\bf{Y}}^{\left( S \right)}} = {\bf{W}}_{{\mathop{\rm out}\nolimits} }^{\left( S \right)}{{\bf{X}}^{\left( S \right)}}
\end{array}
\end{array},} \right.}
\end{array}
\end{equation}
\end{footnotesize}where $S$ is the number of reservoirs and ${{\mathbf{Y}}^{\left( S \right)}}$ is the final output.

\section{Deep recurrent stochastic configuration networks}
\begin{figure}[ht]
	\begin{center}
		\includegraphics[width=9cm]{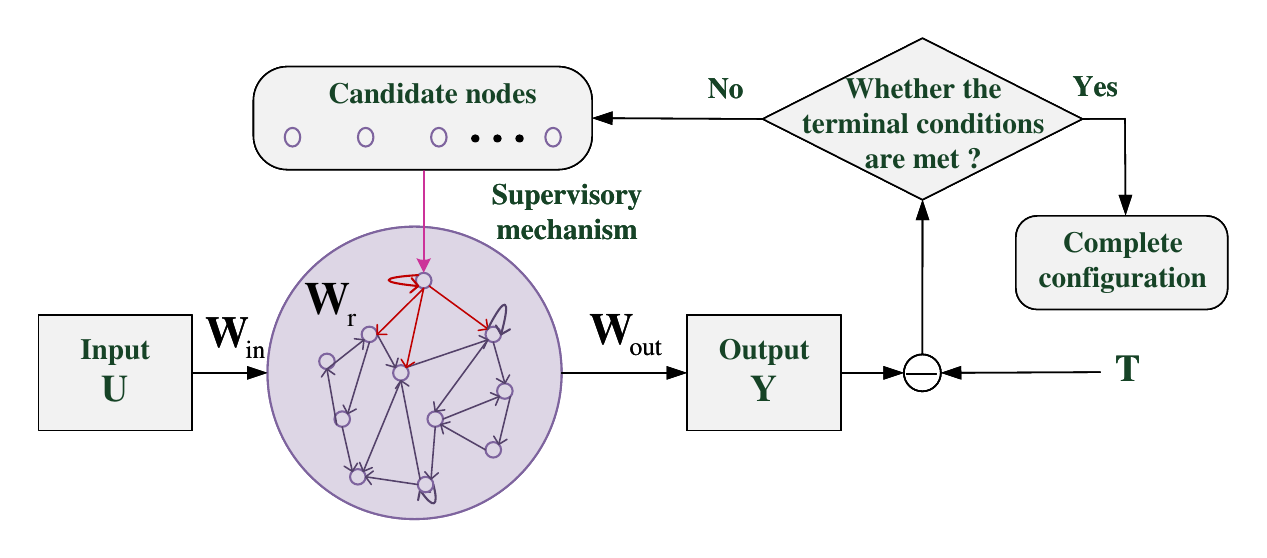}
		\caption{Architecture of the basic RSCN.}
		\label{fig1}
	\end{center}
 \vspace{-0.5cm}
\end{figure}
\begin{figure}[ht]
\vspace{-0.6cm}
	\begin{center}
		\includegraphics[width=9cm]{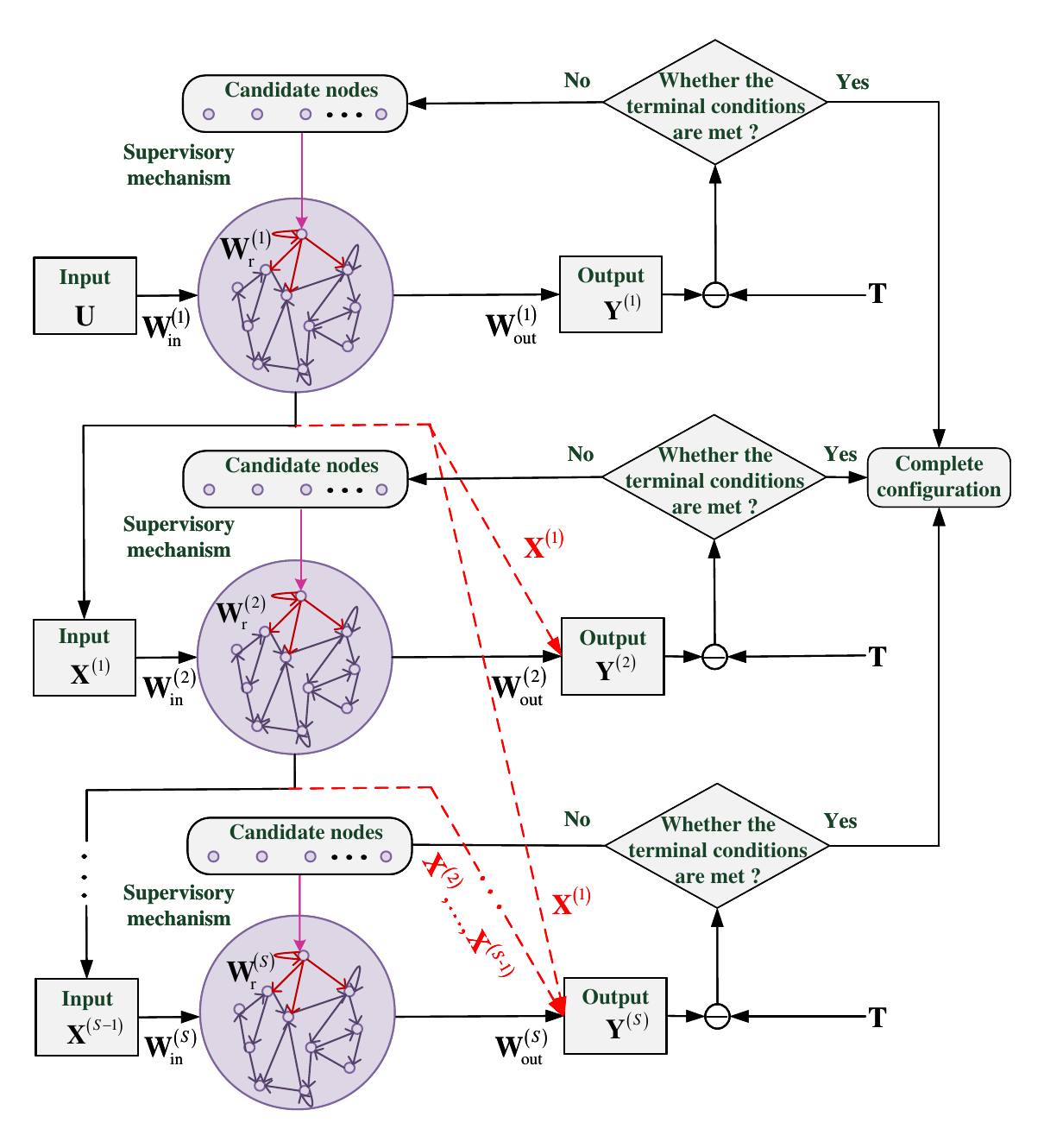}
		\caption{Architecture of the basic DeepRSCN with $S$ reservoir layers.}
		\label{fig2}
	\end{center}
 \vspace{-0.3cm}
\end{figure}

This section extends RSCNs to their deep versions and discusses the fundamentals of DeepRSCNs, including the algorithm description and proof of the universal approximation property. As shown in Fig.~\ref{fig1}, for the original RSCN, a unique reservoir structure is constructed where new node weights are assigned to the primary nodes and itself, while other nodes don't have connections to the new node. DeepRSCNs can be viewed as a stack of multiple RSCNs, with each node being generated through a supervisory mechanism and directly linked to the final output. The architecture of the basic DeepRSCN with $S$ reservoirs is shown in Fig.~\ref{fig2}, and the training algorithm is summarized in Algorithm 1. 
\subsection{Algorithm description}
Given datasets $\left\{ {{\bf{U}},{\bf{T}}} \right\}$, $\mathbf{U}\text{=}\left[ \mathbf{u}(1),\ldots ,\mathbf{u}({{n}_{\max }}) \right]$, $\mathbf{T}=\left[ \mathbf{t}\left( 1 \right),...\mathbf{t}\left( {{n}_{max}} \right) \right]$, construct a DeepRSCN with $S$ layers, and the maximum number of reservoir nodes for each layer is $N_{\max }^{\left( 1 \right)},N_{\max }^{\left( 2 \right)},\ldots ,N_{\max }^{\left( S \right)}$. Assume that each layer has a reservoir with $N$ nodes, that is, \vspace{-0.4cm}

\begin{small}
    \begin{equation} \label{eq11}
\begin{array}{*{20}{c}}
{\left\{ {\begin{array}{*{20}{c}}
{{{\bf{x}}^{\left( 1 \right)}}(n) = g({\bf{W}}_{{\rm{in,}}N}^{\left( 1 \right)}{\bf{u}}(n) + {\bf{W}}_{{\rm{r,}}N}^{\left( 1 \right)}{{\bf{x}}^{\left( 1 \right)}}(n - 1) + {\bf{b}}_N^{\left( 1 \right)})}\\
{{{\bf{Y}}^{\left( 1 \right)}} = {\bf{W}}_{{\mathop{\rm out}\nolimits} }^{\left( 1 \right)}{{\bf{X}}^{\left( 1 \right)}}}
\end{array}} \right.}\\
 \cdots \\
{\left\{ {\begin{array}{*{20}{c}}
{{{\bf{x}}^{\left( j \right)}}(n) = g({\bf{W}}_{{\rm{in,}}N}^{\left( j \right)}{{\bf{X}}^{\left( {j - 1} \right)}} + {\bf{W}}_{{\rm{r,}}N}^{\left( j \right)}{{\bf{x}}^{\left( j \right)}}(n - 1) + {\bf{b}}_N^{\left( j \right)})}\\
{{{\bf{Y}}^{\left( j \right)}} = {\bf{W}}_{{\mathop{\rm out}\nolimits} }^{\left( j \right)}\left[ {\begin{array}{*{20}{c}}
{{{\bf{X}}^{\left( 1 \right)}}}\\
 \vdots \\
{{{\bf{X}}^{\left( j \right)}}}
\end{array}} \right]}
\end{array}} \right.}
\end{array},
\end{equation}
\end{small}where $j=1,2,\ldots ,S$, $\mathbf{W}_{\text{in,}N}^{\left( j \right)}$, $\mathbf{W}_{\text{r,}N}^{\left( j \right)}$, $\mathbf{W}_{\operatorname{out}}^{\left( j \right)}$, $\mathbf{b}_{N}^{\left( j \right)}$, ${{\mathbf{x}}^{\left( j \right)}}(n)$, and ${{\mathbf{Y}}^{\left( j \right)}}$ are the input weight, reservoir weight, output weight, bias, reservoir state, and output of the $j\text{-th}$ layer, respectively, ${{\mathbf{X}}^{\left( j \right)}}\text{=}\left[ {{\mathbf{x}}^{\left( j \right)}}\left( 1 \right),{{\mathbf{x}}^{\left( j \right)}}\left( 2 \right),\ldots ,{{\mathbf{x}}^{\left( j \right)}}\left( {{n}_{\max }} \right) \right]$. Each reservoir is connected to the final output. Let $e_{{{N}_{\text{sum}}}}^{{}}$ be the residual error, where
\begin{equation} \label{eq12}
\left\{ {\begin{array}{*{20}{c}}
{{N_{{\rm{sum}}}} = N,j = 1}\\
{{N_{{\rm{sum}}}} = \sum\limits_{i = 1}^{j - 1} {N_{\max }^{\left( i \right)}}  + N,j \ge 2}
\end{array}} \right..
\end{equation}
When ${{\left\| e_{{{N}_{\text{sum}}}}^{{}} \right\|}_{F}}>\varepsilon $ and Eq. (\ref{eq201}) is not satisfied, we need to add nodes under the supervisory mechanism. Let ${{e}_{0}}=e_{{{N}_{\text{sum}}}}^{{}}$, and the incremental construction process of DeepRSCNs can be summarized as follows.

Step 1: Assign $\left[ w_{\operatorname{i}\operatorname{n},j}^{N+1,1},w_{\operatorname{i}\operatorname{n},j}^{N+1,2},\ldots ,w_{\operatorname{i}\operatorname{n},j}^{N+1,K} \right]$, $b_{N+1}^{j}$, and $\left[ w_{\operatorname{r},j}^{N\text{+1,1}},w_{\operatorname{r},j}^{N\text{+1,2}},\ldots ,w_{\operatorname{r},j}^{N\text{+}1,N+1} \right]$ stochastically in ${{G}_{\max }}$ times from an adjustable uniform distribution $\left[ -\lambda ,\lambda  \right]$. Construct the input weight matrix, bias matrix, and the reservoir weight matrix with a special structure, that is,

\begin{small}
\begin{equation} \label{eq13}
\begin{array}{l}
{\bf{W}}_{{\rm{in}},N + 1}^{\left( j \right)}{\rm{ = }}\left[ {\begin{array}{*{20}{c}}
{w_{{\mathop{\rm i}\nolimits} {\mathop{\rm n}\nolimits} ,j}^{1,1}}&{w_{{\mathop{\rm i}\nolimits} {\mathop{\rm n}\nolimits} ,j}^{1,2}}& \cdots &{w_{{\mathop{\rm i}\nolimits} {\mathop{\rm n}\nolimits} ,j}^{1,K}}\\
{w_{{\mathop{\rm i}\nolimits} {\mathop{\rm n}\nolimits} ,j}^{2,1}}&{w_{{\mathop{\rm i}\nolimits} {\mathop{\rm n}\nolimits} ,j}^{2,2}}& \cdots &{w_{{\mathop{\rm i}\nolimits} {\mathop{\rm n}\nolimits} ,j}^{2,K}}\\
 \vdots & \vdots & \vdots & \vdots \\
{w_{{\mathop{\rm i}\nolimits} {\mathop{\rm n}\nolimits} ,j}^{N,1}}&{w_{{\mathop{\rm i}\nolimits} {\mathop{\rm n}\nolimits} ,j}^{N,2}}& \cdots &{w_{{\mathop{\rm i}\nolimits} {\mathop{\rm n}\nolimits} ,j}^{N,K}}\\
{w_{{\mathop{\rm i}\nolimits} {\mathop{\rm n}\nolimits} ,j}^{N + 1,1}}&{w_{{\mathop{\rm i}\nolimits} {\mathop{\rm n}\nolimits} ,j}^{N + 1,2}}& \cdots &{w_{{\mathop{\rm i}\nolimits} {\mathop{\rm n}\nolimits} ,j}^{N + 1,K}}
\end{array}} \right],\\
{\bf{b}}_{N + 1}^{\left( j \right)} = {\left[ {b_1^j,b_2^j, \ldots ,b_{N + 1}^j} \right]^ \top },\\
{\bf{W}}_{{\rm{r,}}N + 1}^{\left( j \right)}{\rm{ = }}\left[ {\begin{array}{*{20}{c}}
{w_{{\mathop{\rm r}\nolimits} ,j}^{1,1}}&0& \cdots &0&0\\
{w_{{\mathop{\rm r}\nolimits} ,j}^{2,1}}&{w_{{\mathop{\rm r}\nolimits} ,j}^{2,2}}& \cdots &0&0\\
 \vdots & \vdots & \vdots & \vdots & \vdots \\
{w_{{\mathop{\rm r}\nolimits} ,j}^{N,1}}&{w_{{\mathop{\rm r}\nolimits} ,j}^{N,2}}& \cdots &{w_{{\mathop{\rm r}\nolimits} ,j}^{N,N}}&0\\
{w_{{\mathop{\rm r}\nolimits} ,j}^{N{\rm{ + 1,1}}}}&{w_{{\mathop{\rm r}\nolimits} ,j}^{N{\rm{ + 1,2}}}}& \cdots &{w_{{\mathop{\rm r}\nolimits} ,j}^{N{\rm{ + }}1,N}}&{w_{{\mathop{\rm r}\nolimits} ,j}^{N{\rm{ + }}1,N + 1}}
\end{array}} \right].
\end{array}
\end{equation}    
\end{small}

Step 2: Obtain the candidate basis functions $g_{N\text{+}1}^{\left( j \right),1},\ldots ,g_{N\text{+}1}^{\left( j \right),{{G}_{\max }}}$ and substitute them into the following inequality constraint:
\begin{equation} \label{eq14}
\begin{array}{c}
(1 - r - {\mu _{{N_{{\rm{sum}}}} + 1}})\left\| {e_{{N_{{\rm{sum}}}},q}^{}} \right\|^2 - \frac{{\sum\limits_{q = 1}^L {{{\left\langle {e_{{N_{{\rm{sum}}}},q}^{},g_{N{\rm{ + }}1}^{\left( j \right),i}} \right\rangle }^2}} }}{{\left\| {g_{N{\rm{ + }}1}^{\left( j \right),i}} \right\|^2}} \le 0,\\
i = 1,2, \ldots ,{G_{\max }}
\end{array}
\end{equation}
where the non-negative real sequence $\left\{ {{\mu }_{{{N}_{\text{sum}}}+1}} \right\}$ satisfies $\underset{{{N}_{\text{sum}}}\to \infty }{\mathop{\lim }}\,{{\mu }_{{{N}_{\text{sum}}}\text{+}1}}=0$ and ${{\mu }_{{{N}_{\text{sum}}}\text{+}1}}\le \left( 1-r \right)$, $0<r<1$.

Step 3: Find the candidate nodes that satisfy Eq. (\ref{eq14}) and define a set of variables ${{\xi }_{{{N}_{\text{sum}}}}}\text{=}\left[ {{\xi }_{{{N}_{\text{sum}}},1}},...,{{\xi }_{{{N}_{\text{sum}}},L}} \right]$ to select the node making the training error converge as soon as possible,
\begin{equation} \label{eq15}
\begin{array}{l}
{\xi _{N{\rm{ + }}1,q}} = \frac{{{{\left\langle {e_{{N_{{\rm{sum}}}},q}^{},g_{N{\rm{ + }}1}^{\left( j \right)}} \right\rangle }^2}}}{{g_{N{\rm{ + }}1}^{\left( j \right) \top }g_{N{\rm{ + }}1}^{\left( j \right)}}}\\
{\kern 1pt} {\kern 1pt} {\kern 1pt} {\kern 1pt} {\kern 1pt} {\kern 1pt} {\kern 1pt} {\kern 1pt} {\kern 1pt} {\kern 1pt} {\kern 1pt} {\kern 1pt} {\kern 1pt} {\kern 1pt} {\kern 1pt} {\kern 1pt} {\kern 1pt} {\kern 1pt} {\kern 1pt} {\kern 1pt} {\kern 1pt} {\kern 1pt} {\kern 1pt} {\kern 1pt} {\kern 1pt} {\kern 1pt} {\kern 1pt} {\kern 1pt}  - \left( {1 - {\mu _{{N_{{\rm{sum}}}}{\rm{ + }}1}} - r} \right)e_{{N_{{\rm{sum}}}},q}^ \top e_{{N_{{\rm{sum}}}},q}^{}.
\end{array}
\end{equation}
A larger positive value of $\xi _{N}^{{}}\text{=}\sum\limits_{q=1}^{L}{{{\xi }_{N,q}}}$ implies a better configuration of the new node.

Step 4: Update the output weights based on the global least square method, that is,
\begin{equation} \label{eq16}
\begin{array}{l}
{\bf{W}}_{{\mathop{\rm out}\nolimits} }^* = \mathop {\arg \min }\limits_{{\bf{W}}_{{\mathop{\rm out}\nolimits} }^{}} {\left\| {{\bf{T}} - \sum\limits_{z = 1}^j {\sum\limits_{k = 1}^{{N_k}} {{\bf{W}}_{{\mathop{\rm out}\nolimits} ,k}^{\left( z \right)}g_k^{\left( z \right)}} } } \right\|^2},\\
{\kern 1pt} {\kern 1pt} {\kern 1pt} {\kern 1pt} {\kern 1pt} {\kern 1pt} {\kern 1pt} {\kern 1pt} {\kern 1pt} {\kern 1pt} {\kern 1pt} {\kern 1pt} {\kern 1pt} {\kern 1pt} {\kern 1pt} {\kern 1pt} {\kern 1pt} {\kern 1pt} {\kern 1pt} {\kern 1pt} {\kern 1pt} {\kern 1pt} {\kern 1pt} {\kern 1pt} {\kern 1pt} {\kern 1pt} {\kern 1pt} {\kern 1pt} {\kern 1pt} {\kern 1pt} {\kern 1pt} {\kern 1pt} {\kern 1pt} {\kern 1pt} {\kern 1pt} {\kern 1pt} {\kern 1pt} {\kern 1pt} {\kern 1pt} {\kern 1pt} {\kern 1pt} {\kern 1pt} {\kern 1pt} {\kern 1pt} {\kern 1pt} {\kern 1pt} {\kern 1pt} {\kern 1pt} {\kern 1pt} {\kern 1pt} {\kern 1pt} {\kern 1pt} {\kern 1pt} {\kern 1pt} {\kern 1pt} \left\{ {\begin{array}{*{20}{c}}
{z = j,{N_k} = N + 1}\\
{z < j,{N_k} = N_{\max }^{\left( j \right)}}
\end{array}} \right..
\end{array}
\end{equation}

Step 5: Calculate the current residual error $e_{{{N}_{\text{sum}}}+1}^{{}}$, and renew ${{e}_{0}}:=e_{{{N}_{\text{sum}}}+1}^{{}}$, $N=N+1$. If ${{\left\| e_{0}^{{}} \right\|}_{F}}>\varepsilon $ and $N<N_{\max }^{\left( j \right)}$, repeat steps 1–4. If ${{\left\| e_{0}^{{}} \right\|}_{F}}>\varepsilon $ and $N=N_{\max }^{\left( j \right)}$, enter the next layer, update $N=0$, $j=j+1$ and repeat steps 1-4. If ${{\left\| e_{0}^{{}} \right\|}_{F}}\le \varepsilon $ or Eq. (\ref{eq201}) is met or $j=S$ while $N=N_{\max }^{\left( S \right)}$, stop training.

Finally, we have $\underset{{{N}_{\text{sum}}}\to \infty }{\mathop{\lim }}\,\left\| \mathbf{T}-\mathbf{Y}_{{{N}_{\text{sum}}}}^{{}} \right\|=0$, where $\mathbf{Y}_{{{N}_{\text{sum}}}}^{{}}$ is the final readout with ${{N}_{\text{sum}}}$ reservoir nodes.

\subsection{Universal approximation property}
DeepRSCN exploits the supervisory mechanism to construct the reservoir and proceeds to deeper layers, which can approximate any nonlinear maps. This section provides the theoretical result about its universal approximation property.

\textbf{Theorem 2.} Assume span($\Gamma$) is dense on ${{L}_{2}}$ space, for $b_{g}^{*}\in {{\mathbb{R}}^{+}}$, $\forall g\in \Gamma ,0<{{\left\| g \right\|}}<b_{g}^{*}$. Given $0<r<1$ and a non-negative real sequence, $\left\{ {{\mu }_{{{N}_{\text{sum}}}+1}} \right\}$ satisfies $\underset{{{N}_{\text{sum}}}\to \infty }{\mathop{\lim }}\,{{\mu }_{{{N}_{\text{sum}}}+1}}=0$ and ${{\mu }_{{{N}_{\text{sum}}}+1}}\le \left( 1-r \right)$. For ${{N}_{\text{sum}}}=1,2...$, $q=1,2,...,L$, let $e_{{N_{{\rm{sum}}}},q}^*$ denote the residual error and define
\begin{equation} \label{eq17}
\begin{array}{l}
\delta _{{N_{{\rm{sum}}}}{\rm{ + }}1}^* = \sum\limits_{q = 1}^L {\delta _{{N_{{\rm{sum}}}}{\rm{ + }}1,q}^*} {\kern 1pt} ,\\
\delta _{{N_{{\rm{sum}}}}{\rm{ + }}1,q}^* = (1 - r - {\mu _{{N_{{\rm{sum}}}}{\rm{ + }}1}})\left\| {e_{{N_{{\rm{sum}}}},q}^*} \right\|_{}^2.
\end{array}
\end{equation}
If ${{g}_{{{N}_{\text{sum}}}+1}}$ satisfies the following inequality constraint:
\begin{equation} \label{eq18}
{\left\langle {e_{{N_{{\rm{sum}}}},q}^*,g_{{N_{{\rm{sum}}}}{\rm{ + }}1}^{}} \right\rangle ^2} \ge b{_g^*}{^2}\delta _{{N_{{\rm{sum}}}}{\rm{ + }}1,q}^*,
\end{equation}
and the output weights are constructively evaluated by
\begin{equation} \label{eq19}
{\bf{W}}_{{\mathop{\rm out}\nolimits} ,{N_{{\rm{sum}}}}{\rm{ + }}1,q}^* = \frac{{\left\langle {e_{{N_{{\rm{sum}}}},q}^*,g_{{N_{{\rm{sum}}}}{\rm{ + }}1}^{}} \right\rangle }}{{\left\| {g_{{N_{{\rm{sum}}}}{\rm{ + }}1}^{}} \right\|_{}^2}},
\end{equation}
we can obtain $\underset{{{N}_{\text{sum}}}\to \infty }{\mathop{\lim }}\,\left\| \mathbf{T}-\mathbf{Y}_{{{N}_{\text{sum}}}+1}^{{}} \right\|=0$.\\
\textbf{Proof.} With simple computation, we have
\begin{equation} \label{eq20}
\begin{array}{l}
\left\| {e_{{N_{{\rm{sum}}}}{\rm{ + }}1}^*} \right\|_2^2 - \left( {r + {\mu _{{N_{{\rm{sum}}}}{\rm{ + }}1}}} \right)\left\| {e_{{N_{{\rm{sum}}}}}^*} \right\|_2^2\\
{\kern 1pt}  = \sum\limits_{q = 1}^L {\left\langle {e_{{N_{{\rm{sum}}}},q}^* - {\bf{W}}_{{\mathop{\rm out}\nolimits} ,{N_{{\rm{sum}}}} + 1}^*g_{{N_{{\rm{sum}}}} + 1}^{},} \right.} \\
{\kern 1pt} {\kern 1pt} {\kern 1pt} {\kern 1pt} {\kern 1pt} {\kern 1pt} {\kern 1pt} {\kern 1pt} {\kern 1pt} {\kern 1pt} {\kern 1pt} {\kern 1pt} {\kern 1pt} {\kern 1pt} {\kern 1pt} {\kern 1pt} {\kern 1pt} {\kern 1pt} {\kern 1pt} {\kern 1pt} {\kern 1pt} {\kern 1pt} {\kern 1pt} \left. {e_{{N_{{\rm{sum}}}},q}^* - {\bf{W}}_{{\mathop{\rm out}\nolimits} ,{N_{{\rm{sum}}}} + 1}^*g_{{N_{{\rm{sum}}}} + 1}^{}} \right\rangle \\
{\kern 1pt} {\kern 1pt} {\kern 1pt} {\kern 1pt} {\kern 1pt} {\kern 1pt} {\kern 1pt} {\kern 1pt} {\kern 1pt} {\kern 1pt} {\kern 1pt} {\kern 1pt} {\kern 1pt}  - \sum\limits_{q = 1}^L {\left( {r + {\mu _{{N_{{\rm{sum}}}}{\rm{ + }}1}}} \right)\left\langle {e_{{N_{{\rm{sum}}}},q}^*,e_{{N_{{\rm{sum}}}},q}^*} \right\rangle } \\
{\kern 1pt}  = \left( {1 - r - {\mu _{{N_{{\rm{sum}}}}{\rm{ + }}1}}} \right)\left\| {e_{{N_{{\rm{sum}}}},q}^*} \right\|_2^2 - \frac{{\sum\limits_{q = 1}^L {{{\left\langle {e_{{N_{{\rm{sum}}}},q}^*,g_{{N_{{\rm{sum}}}} + 1}^{}} \right\rangle }^2}} }}{{\left\| {g_{{N_{{\rm{sum}}}} + 1}^{}} \right\|_2^2}}\\
 = \delta _{{N_{{\rm{sum}}}}{\rm{ + }}1}^* - \frac{{\sum\limits_{q = 1}^L {{{\left\langle {e_{{N_{{\rm{sum}}}},q}^*,g_{{N_{{\rm{sum}}}} + 1}^{}} \right\rangle }^2}} }}{{\left\| {g_{{N_{{\rm{sum}}}} + 1}^{}} \right\|_2^2}}\\
{\kern 1pt}  \le \delta _{{N_{{\rm{sum}}}}{\rm{ + }}1}^* - \frac{{\sum\limits_{q = 1}^L {{{\left\langle {e_{{N_{{\rm{sum}}}},q}^*,g_{{N_{{\rm{sum}}}} + 1}^{}} \right\rangle }^2}} }}{{b_g^{*2}}} \le 0.
\end{array}
\end{equation}
Then, the following inequalities can be established
\begin{equation} \label{eq21}
\begin{array}{l}
\left\| {e_{{N_{{\rm{sum}}}}{\rm{ + }}1}^*} \right\|_{}^2 \le r\left\| {e_{{N_{{\rm{sum}}}}}^*} \right\|_{}^2 + {\gamma _{{N_{{\rm{sum}}}} + 1}},
\end{array}
\end{equation}
where ${{\gamma _{{N_{{\rm{sum}}}} + 1}} = {\mu _{{N_{{\rm{sum}}}}{\rm{ + }}1}}\left\| {e_{{N_{{\rm{sum}}}}}^*} \right\|_{}^2 \ge 0}$. Note that $\mathop {\lim }\limits_{{N_{{\rm{sum}}}} \to \infty } {\gamma _{{N_{{\rm{sum}}}}}} = 0$. From Eq. (\ref{eq21}), we can easily obtain $\mathop {\lim }\limits_{{N_{{\rm{sum}}}} \to \infty } \left\| {e_{{N_{{\rm{sum}}}}}^*} \right\|_{}^2 = 0$, which completes the proof.

\begin{algorithm}[t]\footnotesize
    \caption{DeepRSC}\label{algo1}	
    \KwIn{Training inputs $\mathbf{U}\text{=}\left[ \mathbf{u}(1),\ldots ,\mathbf{u}({{n}_{\max }}) \right]$, training outputs ${\bf{T}} = {\left[ {{\bf{t}}\left( 1 \right),\ldots ,{\bf{t}}\left( {{n_{max}}} \right)} \right]}$, validation inputs ${{\bf{U}}_{{\rm{val}}}}{\rm{ = }}\left[ {{{\bf{u}}_{{\rm{val}}}}(1),{{\bf{u}}_{{\rm{val}}}}(2), \ldots } \right]$, validation outputs ${{\bf{T}}_{{\rm{val}}}}{\rm{ = }}\left[ {{{\bf{t}}_{{\rm{val}}}}(1),{{\bf{t}}_{{\rm{val}}}}(2), \ldots } \right]$, the maximum number of reservoir layers $S$, the maximum number of reservoir nodes within the $j\text{-th}$ reservoir layer, $N_{\max }^{\left( j \right)}$, $1\le j\le S$, the training error threshold $\varepsilon $, random parameter scalars $\mathbf{\gamma }\text{=}\left\{ {{\lambda }_{1}},{{\lambda }_{2}},...,{{\lambda }_{\max }} \right\}$, the maximum number of stochastic configurations ${{G}_{\max }}$.}
    \KwOut{DeepRSCN model}
    Randomly assign $\mathbf{W}_{\text{in,}N}^{\left( j \right)}$, $\mathbf{b}_{N}^{\left( j \right)}$, and $\mathbf{W}_{\text{r,}N}^{\left( j \right)}$ according to the sparsity of the reservoir from $\left[ -\lambda ,\lambda  \right]$. Calculate the model output $\mathbf{Y}$ and current error $e_{{{N}_{\text{sum}}}}^{{}}$, where $j=1$, and ${{N}_{\text{sum}}}=N$. Set the initial residual error ${{e}_{0}}:=e_{{{N}_{\text{sum}}}}^{{}}$, $0<r<1$, and $\mathbf{\Omega },\mathbf{D}:=\left[ {\kern 1pt} {\kern 1pt} {\kern 1pt} \right]$;\\
    \While{$j \le S$ AND ${\left\| {e_{0}} \right\|_F} > \varepsilon,$}{
        \While{$N < N_{\max }^{\left( S \right)}$ AND ${\left\| {e_{0}} \right\|_F} > \varepsilon,$}{
            \For{$\lambda  \in {\bf{\gamma }}$,}{
                \For{$l=1,2,\ldots ,{{G}_{\max }}$,}{
                    Randomly assign $\mathbf{W}_{\text{in,}N+1}^{\left( j \right)}$, $\mathbf{W}_{\text{r,}N+1}^{\left( j \right)}$, and $\mathbf{b}_{N+1}^{\left( j \right)}$ from $\left[ -\lambda ,\lambda  \right]$ based on Eq. (\ref{eq13});\\
                    Calculate the reservoir state $\mathbf{X}_{{}}^{\left( j \right)}$;\\
                    Calculate ${{\xi }_{{{N}_{\text{sum}}}\text{+}1,q}}$ based on Eq. (\ref{eq15});\\
                    \If{$\min \left\{ {{\xi }_{{{N}_{\text{sum}}}+\text{1,1}}},...,{{\xi }_{{{N}_{\text{sum}}}\text{+1},L}} \right\}\ge 0$}{
                        Save $\mathbf{W}_{\text{in,}N+1}^{\left( j \right)}$, $\mathbf{W}_{\text{r,}N+1}^{\left( j \right)}$, and $\mathbf{b}_{N+1}^{\left( j \right)}$ in $\mathbf{D}$, and $\xi _{{{N}_{\text{sum}}}+1}^{{}}\text{=}\sum\limits_{q=1}^{L}{{{\xi }_{{{N}_{\text{sum}}}+1,q}}}$ in $\mathbf{\Omega }$;\\
                        \Else{go back to Step 5}
                    }                
                }
                \If{$\mathbf{D}$ is not empty}{
                    Find $\mathbf{W}_{\text{in,}N+1}^{\left( j \right)*}$, $\mathbf{W}_{\text{r,}N+1}^{\left( j \right)*}$, and $\mathbf{b}_{N+1}^{\left( j \right)*}$ that maximize ${{\xi }_{{{N}_{\text{sum}}}\text{+}1}}$ in $\mathbf{\Omega }$, and get $\mathbf{X}_{{}}^{\left( j \right)*}$\;
                    \textbf{Break} (go to Step 20)\;
                    \Else{randomly take $\tau \in \left( 0,1-r \right)$ and update $r,r=r+\tau $;}\
                    Return to Step 5;
                }
            }
            Obtain $\mathbf{X}_{{}}^{\left( 1 \right)*}$, $\mathbf{X}_{{}}^{\left( 2 \right)*}$,...,$\mathbf{X}_{{}}^{\left( j \right)*}$, and calculate $\mathbf{W}_{\operatorname{out}}^{*}$ and ${{\mathbf{Y}}^{\left( j \right)*}}$ based on Eq. (\ref{eq16}) and Eq. (\ref{eq11});\\
            Calculate ${{e}_{{{N}_{\text{sum}}}+1}}$ and $e_{{\rm{val,}}{N_{{\rm{sum}}}} + 1}$;\\
            \If{Eq. (\ref{eq201}) is satisfied}{
                Obtain the current network parameters;\\
                \textbf{Break} (go to Step 37)\\
                \Else{Update ${{e}_{0}}:=e_{{{N}_{\text{sum}}}+1}^{{}}$, $N=N+1$, and ${{N}_{\text{sum}}}$ based on Eq. (\ref{eq12});}
            }     
        }
        Set $j=j+1$, $N=0$;    
    }
    \textbf{Return:} $\mathbf{W}_{\operatorname{out}}^{*}$, $\mathbf{W}_{\text{in,}N}^{\left( 1 \right)*},\mathbf{W}_{\text{in,}N}^{\left( 2 \right)*},\ldots ,\mathbf{W}_{\text{in,}N}^{\left( j \right)*}$, $\mathbf{W}_{\text{r,}N}^{\left( 1 \right)*},\mathbf{W}_{\text{r,}N}^{\left( 2 \right)*},\ldots ,\mathbf{W}_{\text{r,}N}^{\left( j \right)*}$, and $\mathbf{b}_{N}^{\left( 1 \right)*},\mathbf{b}_{N}^{\left( 2 \right)*},\ldots ,\mathbf{b}_{N}^{\left( j \right)*}$.
\end{algorithm}

\subsection{Online update of the output weights}
Considering the dynamic characteristics of the process environment, it is essential to adjust learning parameters to effectively handle complex uncertainties. In this subsection, the projection algorithm \cite{ref22} is exploited to update the output weights online in response to the dynamic variations within the systems.

According to Eq. (\ref{eq11}) and Eq. (\ref{eq16}), the model output at $n$ time step can be evaluated by 
\begin{equation} \label{eq22}
{\bf{y}}\left( n \right) = {{\bf{W}}_{{\rm{out}}}}{\bf{G}}\left( n \right),
\end{equation}
where \small ${{\mathbf{W}}_{\text{out}}}=\left[ \mathbf{W}_{\text{out},1}^{\left( 1 \right)},\ldots ,\mathbf{W}_{\text{out},N_{\max }^{\left( 1 \right)}}^{\left( 1 \right)},\ldots ,\mathbf{W}_{\text{out},1}^{\left( S \right)},\ldots ,\mathbf{W}_{\text{out},N}^{\left( S \right)} \right]$, $\mathbf{G}\left( n \right)={{\left[ \mathbf{x}_{1}^{\left( 1 \right)}\left( n \right),\ldots ,\mathbf{x}_{N_{\max }^{\left( 1 \right)}}^{\left( 1 \right)}\left( n \right),\ldots ,\mathbf{x}_{N}^{\left( S \right)}\left( n \right) \right]}^{\top }}$, $\mathbf{W}_{\text{out},k}^{\left( j \right)}$ \normalsize and $\mathbf{x}_{k}^{\left( j \right)}\left( n \right)$ represent the output weight and reservoir state corresponding to the $k\text{-th}$ node in the $j\text{-th}$ layer. Let ${\rm{H}} = \left\{ {{{\bf{W}}_{{\rm{out}}}}:{\bf{y}}(n) = {{\bf{W}}_{{\rm{out}}}}{\bf{G}}\left( n \right)} \right\}$. Choose the closest weight to ${{\mathbf{W}}_{\text{out}}}\left( n-1 \right)$, and define ${{\mathbf{W}}_{\text{out}}}\left( n \right)$ through minimizing the following cost function:
\begin{equation} \label{eq23}
\begin{array}{*{20}{c}}
{J = \frac{1}{2}{{\left\| {{{\bf{W}}_{{\rm{out}}}}\left( n \right) - {{\bf{W}}_{{\rm{out}}}}\left( {n - 1} \right)} \right\|}^2}}\\
{s.t.{\kern 1pt} {\kern 1pt} {\kern 1pt} {\kern 1pt} {\kern 1pt} {\kern 1pt} {\kern 1pt} {\kern 1pt} {\kern 1pt} {\kern 1pt} {\bf{y}}\left( n \right) = {{\bf{W}}_{{\rm{out}}}}\left( n \right){\bf{G}}(n)}
\end{array}.
\end{equation}
By introducing the Lagrange operator ${{\lambda }_{\text{p}}}$, we have
\begin{equation} \label{eq24}
\begin{array}{l}
{J_e} = \frac{1}{2}{\left\| {{{\bf{W}}_{{\rm{out}}}}\left( n \right) - {{\bf{W}}_{{\rm{out}}}}\left( {n - 1} \right)} \right\|^2}\\
{\kern 1pt} {\kern 1pt} {\kern 1pt} {\kern 1pt} {\kern 1pt} {\kern 1pt} {\kern 1pt} {\kern 1pt} {\kern 1pt} {\kern 1pt} {\kern 1pt} {\kern 1pt} {\kern 1pt} {\kern 1pt}  + {\lambda _{\rm{p}}}\left( {{\bf{y}}\left( n \right) - {{\bf{W}}_{{\rm{out}}}}\left( n \right){\bf{G}}\left( n \right)} \right).
\end{array}
\end{equation}
The necessary conditions for ${{J}_{e}}$ to be minimal are
\begin{equation} \label{eq25}
\left\{ {\begin{array}{*{20}{c}}
{\frac{{\partial {J_e}}}{{\partial {{\bf{W}}_{{\rm{out}}}}\left( n \right)}} = 0}\\
{\frac{{\partial {J_e}}}{{\partial {\lambda _{\rm{p}}}}} = 0}
\end{array}} \right..
\end{equation}
Then, we can obtain
\begin{equation} \label{eq26}
\left\{ {\begin{array}{*{20}{c}}
{{{\bf{W}}_{{\rm{out}}}}\left( n \right) - {{\bf{W}}_{{\rm{out}}}}\left( {n - 1} \right) - {\lambda _{\rm{p}}}{{\bf{G}}^ \top }(n) = 0}\\
{{\bf{y}}\left( n \right) - {{\bf{W}}_{{\rm{out}}}}\left( n \right){\bf{G}}\left( n \right) = 0}
\end{array}} \right.,
\end{equation}
\begin{equation} \label{eq27}
{\lambda _{\rm{p}}} = \frac{{{\bf{y}}\left( n \right) - {{{\bf{W}}}_{{\rm{out}}}}\left( {n - 1} \right){\bf{\hat g}}(n)}}{{{\bf{\hat g}}{{(n)}^ \top }{\bf{\hat g}}(n)}}.
\end{equation}
Substituting Eq. (\ref{eq27}) into Eq. (\ref{eq26}), yields 
\begin{equation} \label{eq28}
\begin{array}{l}
{{\bf{W}}_{{\rm{out}}}}(n) = {{\bf{W}}_{{\rm{out}}}}(n - 1) + \\
{\kern 1pt} \frac{{{\bf{G}}{{(n)}^ \top }}}{{{\bf{G}}{{(n)}^ \top }{\bf{G}}(n)}}\left( {{\bf{y}}\left( n \right) - {{\bf{W}}_{{\rm{out}}}}\left( {n - 1} \right){\bf{G}}(n)} \right).
\end{array}
\end{equation}
Particularly, a small constant $c$ is added to the denominator to avoid division by zero and a coefficient $a$ is multiplied by the numerator to obtain the improved projection algorithm, that is,
\begin{equation} \label{eq29}
\begin{array}{l}
{{\bf{W}}_{{\rm{out}}}}(n) = {{\bf{W}}_{{\rm{out}}}}(n - 1) + \\
\frac{{a{\bf{G}}{{(n)}^ \top }}}{{c + {\bf{G}}{{(n)}^ \top }{\bf{G}}(n)}}\left( {{\bf{y}}\left( n \right) - {{\bf{W}}_{{\rm{out}}}}\left( {n - 1} \right){\bf{G}}(n)} \right),
\end{array}
\end{equation}
where $0<a\le 1$ and $c>0$.
\begin{remark}
We have conducted the stability and convergence analysis of the proposed approach based on the projection algorithm and introduced an enhanced condition to further improve the model’s stability. For more details, one can refer to \cite{ref181}.  
\end{remark}
\section{Experiment results}
In this section, the performance of the DeepRSCN is tested on the Mackey-Glass time series prediction, a nonlinear system identification task, and two industrial data predictive analyses. The performance comparisons are conducted between the original ESN, DeepESN, RSCN and our proposed DeepRSCN. For simplicity, DeepESNs with two and three layers are denoted as DeepESN2 and DeepESN3, while DeepRSCNs with two and three layers are denoted as DeepRSCN2 and DeepRSCN3. The normalized root means square error (NRMSE) is used to evaluate the model performance, that is,
\begin{equation} \label{eq31}
NRMSE = \sqrt {\frac{{\sum\limits_{n = 1}^{{n_{max}}} {{{\left( {{\bf{y}}\left( n \right) - {\bf{t}}\left( n \right)} \right)}^2}} }}{{{n_{max}}{\sigma ^2}}}} ,
\end{equation}
where ${{\sigma }^{2}}$ denotes the variance of the desired output $\mathbf{t}$.

The following key parameters are taken as: the scaling factor of spectral radius $\alpha  \in \left[ {0.5,1} \right]$, and the sparsity of the reservoir weight ranges from 0.01 to 0.03. For the ESN and DeepESNs, the input ${{\mathbf{W}}_{\text{in}}}$ and reservoir weights ${{\mathbf{W}}_{\text{r}}}$ are randomly generated from a uniform distribution $\left[ -1,1 \right]$. For the RSC-based frameworks, the maximum number of stochastic configurations is set to ${{G}_{\max }}=100$, weight scale sequence $\left\{ 0.5,1,5,10,30,50,100 \right\}$, contractive sequence $r=\left[ 0.9,0.99,0.999,0.9999 \right]$, training tolerance ${{\varepsilon }}={{10}^{-6}}$, the step size ${{N}_{\text{step}}}=6$, and the initial reservoir size is set to 5. Specifically, the grid search method is used to determine the hyperparameters based on the validation error. Each simulation is performed using 30 independent trials under similar conditions, and the mean and standard deviation of NRMSE and training time are utilized for evaluation.

\begin{table*}[]
\caption{Performance comparison of different models on MG tasks.} \label{tb1}
\centering
\begin{tabular}{cccccc}
\hline
Datasets             & Models     & Reservoir size & Training time (s)           & Training NRMSE           & Testing NRMSE                       \\ \hline
\multirow{6}{*}{MG}  & ESN        & 96             & 0.14193±0.03365          & 0.01125±0.00233          & 0.02618±0.00776                     \\
                     & DeepESN2   & 72-13          & 0.09147±0.01701          & 0.01018±0.00303          & 0.02301±0.00302                     \\
                     & DeepESN3   & 48-19-12       & \textbf{0.04536±0.02058} & 0.00865±0.00175          & 0.01891±0.00281                     \\
                     & RSCN       & 67             & 0.95109±0.08382          & 0.00342±0.00041          & 0.01206±0.00558                     \\
                     & DeepRSCN2 & 40-22          & 0.88100±0.07235          & 0.00310±0.00083          & 0.00921±0.00080                     \\
                     & DeepRSCN3  & 25-25-8        & 0.84581±0.07998          & \textbf{0.00297±0.00064} & \textbf{0.00888±0.00184}            \\ \hline
\multirow{6}{*}{MG1} & ESN        & 124            & 0.16983±0.05084          & 0.01572±0.00679          & 0.03983±0.01130                     \\
                     & DeepESN2   & 78-25          & 0.09932±0.01526          & 0.01274±0.00334          & \multicolumn{1}{l}{0.03146±0.00299} \\
                     & DeepESN3   & 46-39-17       & \textbf{0.09515±0.01913} & 0.01124±0.00199          & 0.02943±0.00152                     \\
                     & RSCN       & 79             & 0.88928±0.57240          & 0.00511±0.00032          & 0.01467±0.00353                     \\
                     & DeepRSCN2 & 50-16          & 0.61302±0.12173          & \textbf{0.00487±0.00029}          & 0.01237±0.00114                     \\
                     & DeepRSCN3  & 30-30-9        & 0.61169±0.11058          & 0.00491±0.00036 & \textbf{0.01144±0.00223}            \\ \hline
\multirow{6}{*}{MG2} & ESN        & 135            & 0.15339±0.07362          & 0.03091±0.01927          & 0.08309±0.01122                     \\
                     & DeepESN2   & 73-65          & 0.11967±0.01442          & 0.02642±0.00583          & 0.06465±0.00587                     \\
                     & DeepESN3   & 56-52-23       & \textbf{0.11564±0.00617} & 0.02371±0.00393          & 0.06078±0.00374                     \\
                     & RSCN       & 105             & 1.19886±0.28293          & 0.00719±0.00589          & 0.03431±0.00278                     \\
                     & DeepRSCN2 & 60-35          & 0.91460±0.12470          & 0.00570±0.00293          & 0.03122±0.00259                     \\
                     & DeepRSCN3  & 40-40-11       & 0.89239±0.14405          & \textbf{0.00552±0.00424} & \textbf{0.02845±0.00356}            \\ \hline
\end{tabular}
\vspace{-0.3cm}
\end{table*}

\begin{figure*}[htbp]
\vspace{-0.2cm}
	\centering
	\subfloat{\includegraphics[width=7cm]{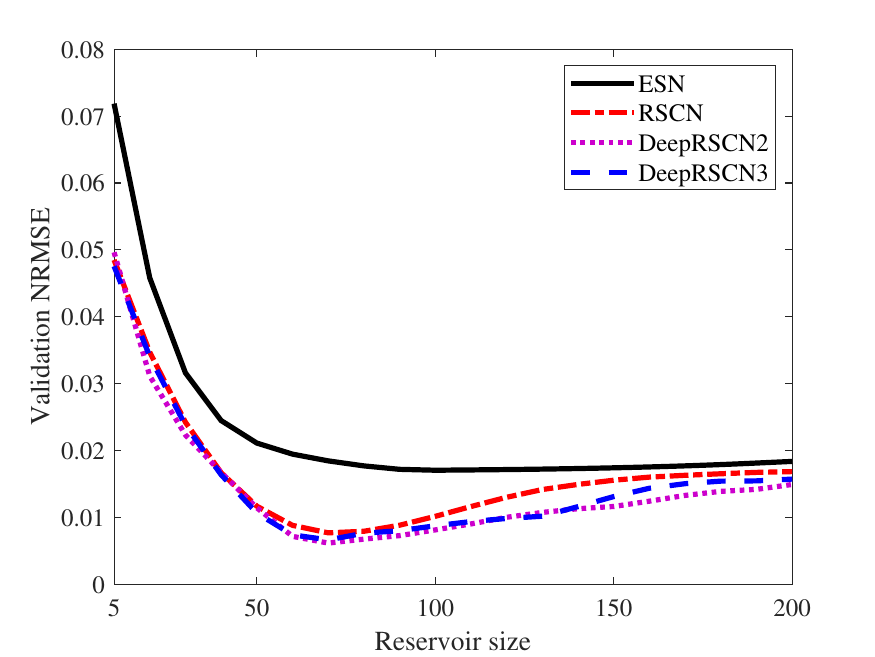}}
	\subfloat{\includegraphics[width=7cm]{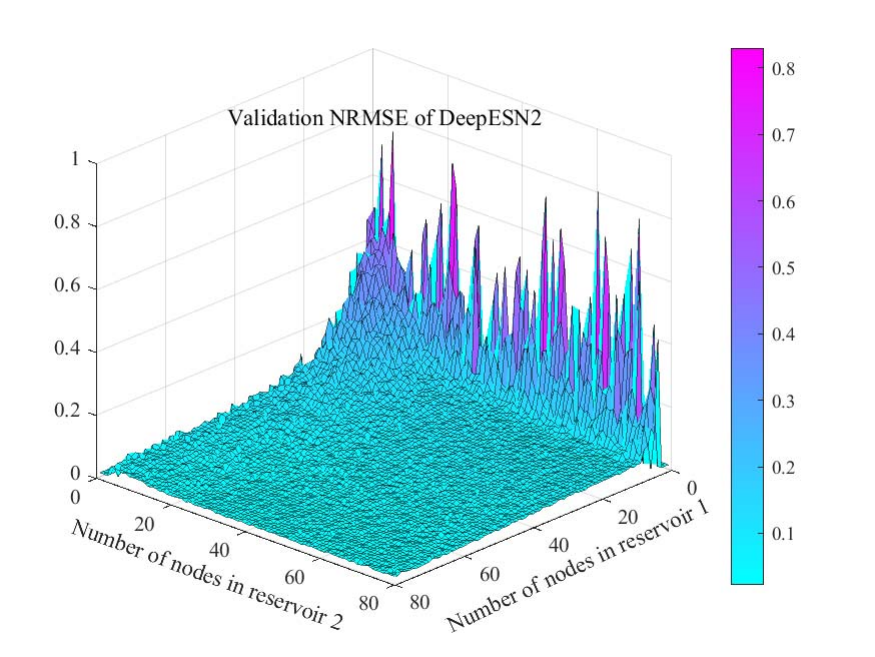}}
	\caption{Validation performance of different models on the MG task.}
	\label{fig3}
 \vspace{-0.5cm}
\end{figure*}
\subsection{Mackey–Glass system (MGS)}
MGS is often used for time series prediction and nonlinear system identification due to its chaotic behavior. MGS can be represented by the following differential equation with a time delay, that is,
\begin{equation} \label{eq32}
\frac{{dy}}{{dn}} = \upsilon y\left( n \right) + \frac{{\alpha y\left( {n - \tau } \right)}}{{1 + y{{\left( {n - \tau } \right)}^{10}}}}.
\end{equation}
When $\tau >16.8$, the entire sequence is chaotic, acyclic, nonconvergent, and divergent. Generally, the parameters in Eq. (\ref{eq32}) are set to $\upsilon \text{=}-0.1$, $\alpha \text{=}0.2$, and $\tau =17$. The initial values $\left\{ {y\left( 0 \right), \ldots ,y\left( \tau  \right)} \right\}$ are generated from $\left[ {0.1,1.3} \right]$. $\left\{ {y\left( n \right),y\left( {n - 6} \right),y\left( {n - 12} \right),y\left( {n - 18} \right)} \right\}$, as the model inputs, are used to predict $y\left( {n + 6} \right)$. and the second-order Runge–Kutta method is used to generate 1177 sequence points. In our simulation, samples of 1–500 steps are selected to train the network, 501–800 steps are used as the validation data, and the remaining samples are used for testing. The first 20 samples of each set are washed out. Considering the order uncertainty,  we assume that certain orders remain unknown and design two experimental setups. In the MG1 task, we select ${{\mathbf{u}}}\left( n \right)={{\left[ y\left( n-6  \right),y\left( n-12  \right),y\left( n-18  \right) \right]}}$ to predict $y\left( {n + 6} \right)$. In the MG2 task, the input is set as ${{\mathbf{u}}}\left( n \right)={{\left[ y\left( n-12  \right),y\left( n-18  \right) \right]}^{\top }}$. These configurations are intentionally designed to evaluate the effectiveness of RSCN with incomplete input variables.

It is well-known that the reservoir size has a significant impact on the model performance. To determine the optimal network structure, we observe the validation NRMSE curves. For fairness, an equal number of reservoir nodes are allocated to both single-layer and multi-layer frameworks. Fig.~\ref{fig3} depicts the validation performance of various models on the MG task. The validation NRMSE gradually decreases with increasing reservoir sizes, indicating that the model is underfitted. However, as the number of nodes increasing, the validation NRMSE increases beyond the minimum point, indicating that the model is overfitted. Evidently, the optimal reservoir sizes for ESN, DeepESN2, RSCN, DeepRSCN2, and DeepRSCN3 are determined to be 96, 72–33, 54, 75–14, and 50–50–5, respectively. Furthermore, the validation performance of the recurrent stochastic configuration (RSC) frameworks is superior to that of ESN, indicating that RSC-based models can contribute to sound performance if certain tricks are adopted to prevent overfitting.

Table~\ref{tb1} lists a comprehensive performance comparison of different models on the three MG tasks. It can be seen that the NRMSE of different models gradually increases on the MG-MG2 tasks, indicating that each significant input variable contributes to the final output. DeepRSCN3 demonstrates better performance compared to other models in both training (excluding the MG1 task) and testing sets. Moreover, compared with ESN and its deep variants, RSC-based methods have a smaller reservoir size and lower training and testing NRMSE. Thus, they can offer more compact reservoirs and minimize the influence of order uncertainty. The comparison between the RSCN and DeepRSCNs reveals that deeper reservoir frameworks can lead to superior performance, verifying the effectiveness of the proposed approach.
\begin{figure*}[htbp] 
\vspace{-0.3cm}
	\centering
	\subfloat[MG]{\includegraphics[width=5.5cm]{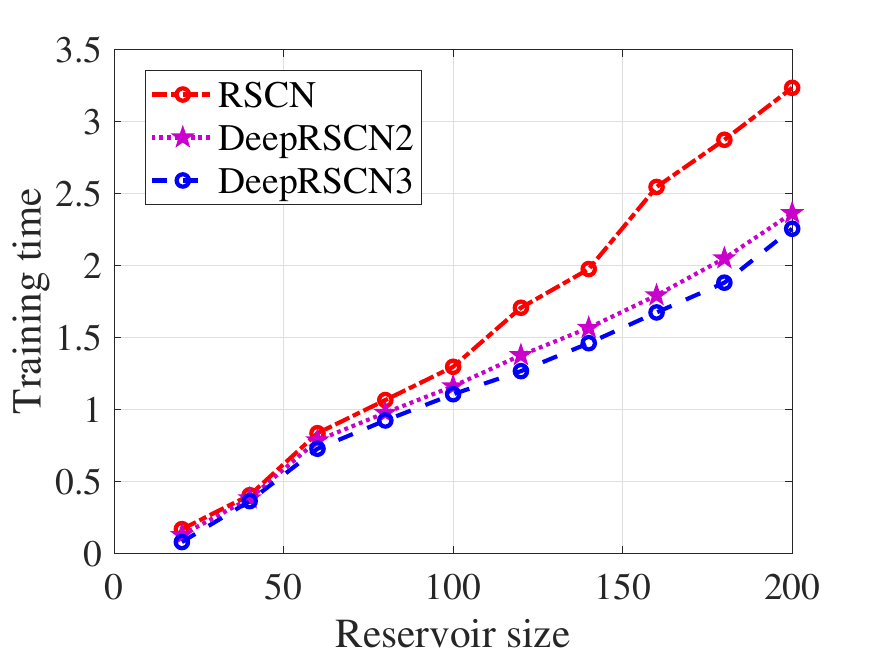}}
	\subfloat[MG1]{\includegraphics[width=5.5cm]{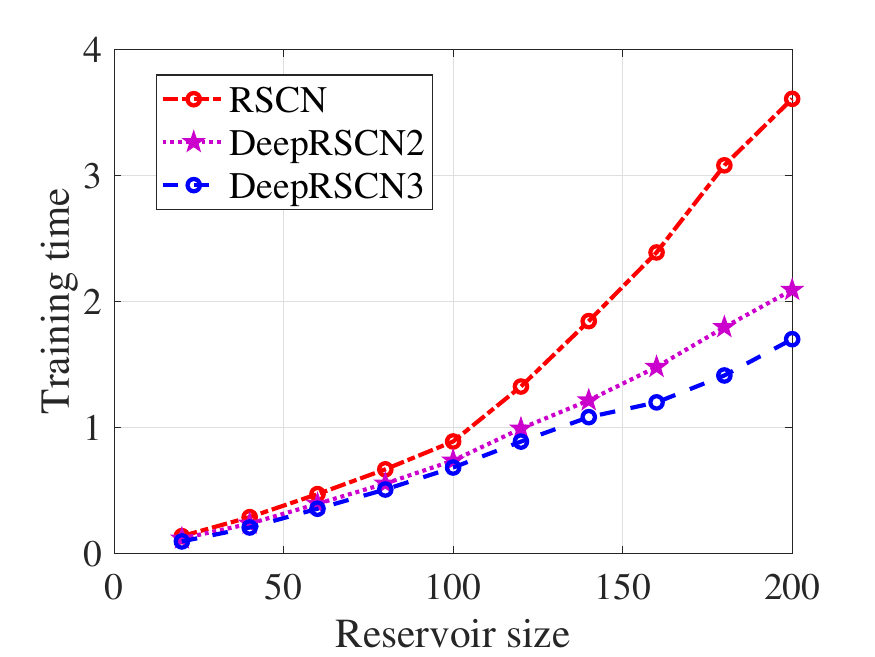}}
        \subfloat[MG2]{\includegraphics[width=5.5cm]{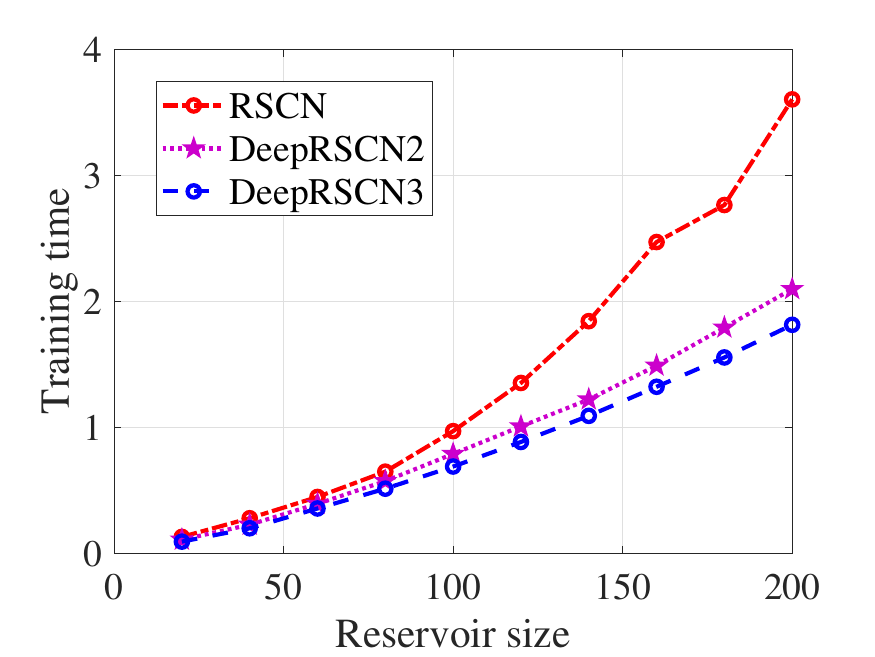}}
	\caption{Training time comparison between the RSC-based frameworks with different reservoir sizes on the three MG tasks. (The reservoir size represents the total number of nodes across all layers, with an even number of nodes in each layer.)}
	\label{fig4}
 \vspace{-0.5cm}
\end{figure*}
\setlength{\abovecaptionskip}{-0.5cm}
\begin{figure}[htbp] 
\vspace{-0.5cm}
	\centering
        \includegraphics[width=7cm]{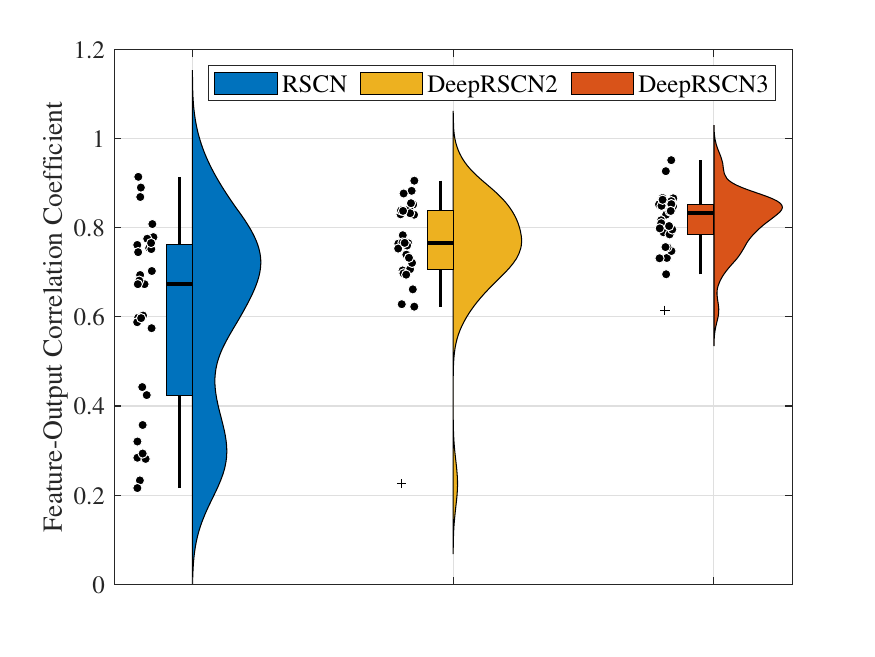}
	\caption{The correlation between reservoir outputs and target outputs across varying model depths on the nonlinear system identification task.}
	\label{fig511}
 \vspace{-0.3cm}
\end{figure}

Fig.~\ref{fig4} illustrates the training time of the RSC-based networks for different reservoir sizes across three MG tasks. DeepRSCN3 exhibits a shorter training time on all tasks, particularly for large reservoir sizes. This indicates that our proposed DeepRSCNs outperform the original RSCN in terms of computational efficiency.\vspace{-0.3cm}
\subsection{Nonlinear system identification}
The nonlinear dynamic system can be expressed as follows:
\begin{equation} \label{eq33}
\begin{array}{l}
y\left( {n + 1} \right) = 0.72y\left( n \right) + 0.025y\left( {n - 1} \right)u\left( {n - 1} \right)\\
{\kern 1pt} {\kern 1pt} {\kern 1pt} {\kern 1pt} {\kern 1pt} {\kern 1pt} {\kern 1pt} {\kern 1pt} {\kern 1pt} {\kern 1pt} {\kern 1pt} {\kern 1pt} {\kern 1pt} {\kern 1pt} {\kern 1pt} {\kern 1pt} {\kern 1pt} {\kern 1pt} {\kern 1pt} {\kern 1pt} {\kern 1pt} {\kern 1pt} {\kern 1pt} {\kern 1pt} {\kern 1pt} {\kern 1pt} {\kern 1pt} {\kern 1pt} {\kern 1pt} {\kern 1pt} {\kern 1pt} {\kern 1pt} {\kern 1pt} {\kern 1pt} {\kern 1pt} {\kern 1pt} {\kern 1pt} {\kern 1pt} {\kern 1pt} {\kern 1pt} {\kern 1pt} {\kern 1pt} {\kern 1pt} {\kern 1pt} {\kern 1pt} {\kern 1pt} {\kern 1pt} {\kern 1pt} {\kern 1pt} {\kern 1pt} {\kern 1pt} {\kern 1pt}  + 0.01{u^2}\left( {n - 2} \right) + 0.2u\left( {n - 3} \right).
\end{array}
\end{equation}
In the training stage, $u\left( n \right)$ is generated from the uniform distribution $\left[ { - 1,1} \right]$ and the initial output values are $y\left( 1 \right)=y\left( 2 \right)=y\left( 3 \right)=0, y\left( 4 \right)=0.1.$ In the testing stage, the input is generated by
\begin{equation} \label{eq34}
u\left( n \right) = \left\{ {\begin{array}{*{20}{l}}
{\sin \left( {\frac{{\pi n}}{{25}}} \right),{\kern 1pt} {\kern 1pt} {\kern 1pt} {\kern 1pt} {\kern 1pt} {\kern 1pt} {\kern 1pt} {\kern 1pt} {\kern 1pt} {\kern 1pt} {\kern 1pt} {\kern 1pt} {\kern 1pt} {\kern 1pt} {\kern 1pt} {\kern 1pt} {\kern 1pt} {\kern 1pt} {\kern 1pt} {\kern 1pt} {\kern 1pt} {\kern 1pt} {\kern 1pt} {\kern 1pt} {\kern 1pt} {\kern 1pt} {\kern 1pt} {\kern 1pt} {\kern 1pt} {\kern 1pt} {\kern 1pt} {\kern 1pt} {\kern 1pt} {\kern 1pt} {\kern 1pt} {\kern 1pt} {\kern 1pt} {\kern 1pt} {\kern 1pt} {\kern 1pt} {\kern 1pt} {\kern 1pt} {\kern 1pt} {\kern 1pt} {\kern 1pt} {\kern 1pt} {\kern 1pt} {\kern 1pt} {\kern 1pt} {\kern 1pt} {\kern 1pt} {\kern 1pt} {\kern 1pt} {\kern 1pt} {\kern 1pt} {\kern 1pt} {\kern 1pt} {\kern 1pt} {\kern 1pt} {\kern 1pt} {\kern 1pt} {\kern 1pt} {\kern 1pt} {\kern 1pt} {\kern 1pt} {\kern 1pt} {\kern 1pt} {\kern 1pt} {\kern 1pt} {\kern 1pt} {\kern 1pt} {\kern 1pt} {\kern 1pt} {\kern 1pt} {\kern 1pt} {\kern 1pt} {\kern 1pt} {\kern 1pt} {\kern 1pt} {\kern 1pt} {\kern 1pt} {\kern 1pt} {\kern 1pt} {\kern 1pt} {\kern 1pt} {\kern 1pt} {\kern 1pt} {\kern 1pt} {\kern 1pt} {\kern 1pt} {\kern 1pt} {\kern 1pt} {\kern 1pt} {\kern 1pt} {\kern 1pt} {\kern 1pt} {\kern 1pt} {\kern 1pt} {\kern 1pt} {\kern 1pt} {\kern 1pt} {\kern 1pt} 0 < n < 250}\\
{1,{\kern 1pt}  {\kern 1pt} {\kern 1pt} {\kern 1pt} {\kern 1pt} {\kern 1pt} {\kern 1pt} {\kern 1pt} {\kern 1pt} {\kern 1pt} {\kern 1pt} {\kern 1pt} {\kern 1pt} {\kern 1pt} {\kern 1pt} {\kern 1pt} {\kern 1pt} {\kern 1pt} {\kern 1pt} {\kern 1pt} {\kern 1pt} {\kern 1pt} {\kern 1pt} {\kern 1pt} {\kern 1pt} {\kern 1pt} {\kern 1pt} {\kern 1pt} {\kern 1pt} {\kern 1pt} {\kern 1pt} {\kern 1pt} {\kern 1pt} {\kern 1pt} {\kern 1pt} {\kern 1pt} {\kern 1pt} {\kern 1pt} {\kern 1pt} {\kern 1pt} {\kern 1pt} {\kern 1pt} {\kern 1pt} {\kern 1pt} {\kern 1pt} {\kern 1pt} {\kern 1pt} {\kern 1pt} {\kern 1pt} {\kern 1pt} {\kern 1pt} {\kern 1pt} {\kern 1pt} {\kern 1pt} {\kern 1pt} {\kern 1pt}{\kern 1pt} {\kern 1pt} {\kern 1pt} {\kern 1pt} {\kern 1pt} {\kern 1pt} {\kern 1pt} {\kern 1pt} {\kern 1pt} {\kern 1pt} {\kern 1pt} {\kern 1pt} {\kern 1pt} {\kern 1pt} {\kern 1pt} {\kern 1pt} {\kern 1pt} {\kern 1pt} {\kern 1pt} {\kern 1pt} {\kern 1pt} {\kern 1pt} {\kern 1pt} {\kern 1pt} {\kern 1pt} {\kern 1pt} {\kern 1pt} {\kern 1pt} {\kern 1pt} {\kern 1pt} {\kern 1pt} {\kern 1pt} {\kern 1pt} {\kern 1pt} {\kern 1pt} {\kern 1pt} {\kern 1pt} {\kern 1pt} {\kern 1pt} {\kern 1pt} {\kern 1pt} {\kern 1pt} {\kern 1pt} {\kern 1pt} {\kern 1pt} {\kern 1pt} {\kern 1pt} {\kern 1pt} {\kern 1pt} {\kern 1pt} {\kern 1pt} {\kern 1pt} {\kern 1pt} {\kern 1pt} {\kern 1pt} {\kern 1pt} {\kern 1pt} {\kern 1pt} {\kern 1pt} {\kern 1pt} {\kern 1pt} {\kern 1pt} {\kern 1pt} {\kern 1pt} {\kern 1pt} {\kern 1pt} {\kern 1pt} {\kern 1pt} {\kern 1pt} {\kern 1pt} 250 \le n < 500}\\
{ - 1,{\kern 1pt}  {\kern 1pt} {\kern 1pt}  {\kern 1pt} {\kern 1pt} {\kern 1pt} {\kern 1pt} {\kern 1pt} {\kern 1pt} {\kern 1pt}{\kern 1pt} {\kern 1pt} {\kern 1pt} {\kern 1pt} {\kern 1pt} {\kern 1pt} {\kern 1pt} {\kern 1pt} {\kern 1pt} {\kern 1pt} {\kern 1pt} {\kern 1pt} {\kern 1pt} {\kern 1pt} {\kern 1pt} {\kern 1pt} {\kern 1pt} {\kern 1pt} {\kern 1pt} {\kern 1pt} {\kern 1pt} {\kern 1pt} {\kern 1pt} {\kern 1pt} {\kern 1pt} {\kern 1pt} {\kern 1pt} {\kern 1pt} {\kern 1pt} {\kern 1pt} {\kern 1pt} {\kern 1pt} {\kern 1pt} {\kern 1pt} {\kern 1pt} {\kern 1pt} {\kern 1pt} {\kern 1pt} {\kern 1pt} {\kern 1pt} {\kern 1pt} {\kern 1pt} {\kern 1pt} {\kern 1pt} {\kern 1pt} {\kern 1pt} {\kern 1pt} {\kern 1pt} {\kern 1pt} {\kern 1pt} {\kern 1pt} {\kern 1pt} {\kern 1pt} {\kern 1pt} {\kern 1pt} {\kern 1pt} {\kern 1pt} {\kern 1pt} {\kern 1pt} {\kern 1pt} {\kern 1pt} {\kern 1pt} {\kern 1pt} {\kern 1pt} {\kern 1pt} {\kern 1pt} {\kern 1pt} {\kern 1pt} {\kern 1pt} {\kern 1pt} {\kern 1pt} {\kern 1pt} {\kern 1pt} {\kern 1pt} {\kern 1pt} {\kern 1pt} {\kern 1pt} {\kern 1pt} {\kern 1pt} {\kern 1pt} {\kern 1pt} {\kern 1pt} {\kern 1pt} {\kern 1pt} {\kern 1pt} {\kern 1pt} {\kern 1pt} {\kern 1pt} {\kern 1pt} {\kern 1pt} {\kern 1pt} {\kern 1pt} {\kern 1pt} {\kern 1pt} {\kern 1pt} {\kern 1pt} {\kern 1pt} {\kern 1pt} {\kern 1pt} {\kern 1pt} {\kern 1pt} {\kern 1pt} {\kern 1pt} {\kern 1pt} {\kern 1pt} {\kern 1pt} {\kern 1pt} {\kern 1pt} 500 \le n < 750}\\
\begin{array}{l}
0.6\cos \left( {\frac{{\pi n}}{{10}}} \right) + 0.1\cos \left( {\frac{{\pi n}}{{32}}} \right) + \\
{\kern 1pt} {\kern 1pt} {\kern 1pt} {\kern 1pt} {\kern 1pt} 0.3\sin \left( {\frac{{\pi n}}{{25}}} \right),{\kern 1pt} {\kern 1pt} {\kern 1pt} {\kern 1pt} {\kern 1pt} {\kern 1pt} {\kern 1pt} {\kern 1pt} {\kern 1pt} {\kern 1pt} {\kern 1pt} {\kern 1pt} {\kern 1pt}  {\kern 1pt} {\kern 1pt} {\kern 1pt} {\kern 1pt} {\kern 1pt} {\kern 1pt} {\kern 1pt} {\kern 1pt} {\kern 1pt} {\kern 1pt} {\kern 1pt} {\kern 1pt} {\kern 1pt} {\kern 1pt} {\kern 1pt} {\kern 1pt} {\kern 1pt} {\kern 1pt} {\kern 1pt} {\kern 1pt} {\kern 1pt} {\kern 1pt} {\kern 1pt} {\kern 1pt} {\kern 1pt} {\kern 1pt} {\kern 1pt} {\kern 1pt} {\kern 1pt} {\kern 1pt} {\kern 1pt} {\kern 1pt} {\kern 1pt} {\kern 1pt} {\kern 1pt} {\kern 1pt} {\kern 1pt} {\kern 1pt} {\kern 1pt} {\kern 1pt} {\kern 1pt} {\kern 1pt} {\kern 1pt} {\kern 1pt} {\kern 1pt} {\kern 1pt} {\kern 1pt} {\kern 1pt} {\kern 1pt} {\kern 1pt} {\kern 1pt} {\kern 1pt} {\kern 1pt} {\kern 1pt} 750 \le n \le 1000.
\end{array}
\end{array}} \right.
\end{equation}
$y\left( n \right)$ and $u\left( n \right)$ are used to predict $y\left( n+1 \right)$. This dataset consists of 2000 samples for training, 1000 samples for validation, and 1000 samples for testing. The first 100 samples of each set are washed out.
\setlength{\abovecaptionskip}{5pt}
\begin{table*}[]
\caption{Performance comparison of different models on the nonlinear system identification task.} \label{tb2}
\centering
\begin{tabular}{ccccc}
\hline
Models     & Reservoir size & Training time (s)            & Training NRMSE           & Testing NRMSE            \\ \hline
ESN        & 157            & 0.06339±0.01033          & 0.00916±0.00252          & 0.06276±0.00325          \\
DeepESN2   & 59-45          & 0.02522±0.00240          & 0.00592±0.00064          & 0.05631±0.00763          \\
DeepESN3   & 37-33-28       & \textbf{0.02262±0.00319} & 0.00580±0.00104          & 0.04221±0.00821          \\
RSCN       & 102            & 2.23162±0.85353         & 0.00794±0.00057          & 0.03958±0.00097          \\
DeepRSCN2 & 60-18          & 1.64534±0.07554          & 0.00579±0.00148          & 0.02835±0.00706          \\
DeepRSCN3  & 40-40-5        & 1.65239±0.29147          & \textbf{0.00534±0.00152} & \textbf{0.02583±0.00822} \\ \hline
\end{tabular}
\vspace{-0.5cm}
\end{table*}

\begin{figure*}[htbp]
\vspace{-0.3cm}
	\centering
	\subfloat[ESN]{\includegraphics[width=5.5cm]{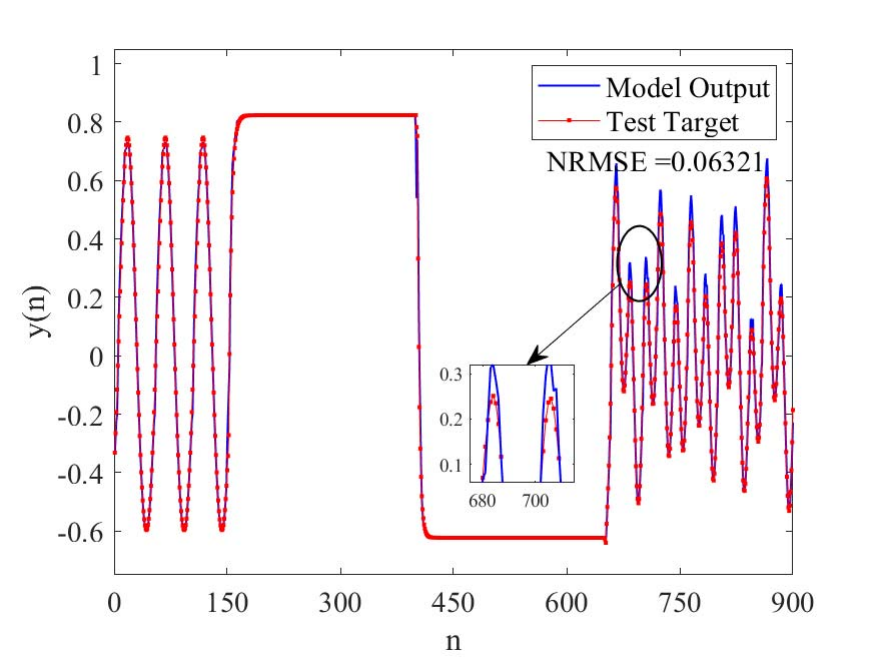}}
	\subfloat[DeepESN2]{\includegraphics[width=5.5cm]{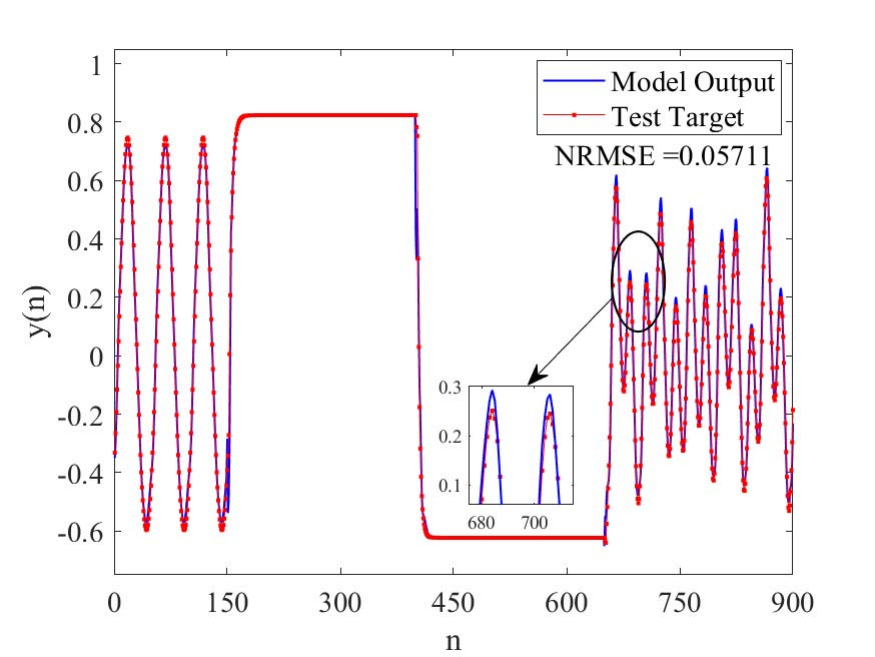}}
    \subfloat[DeepESN3]{\includegraphics[width=5.5cm]{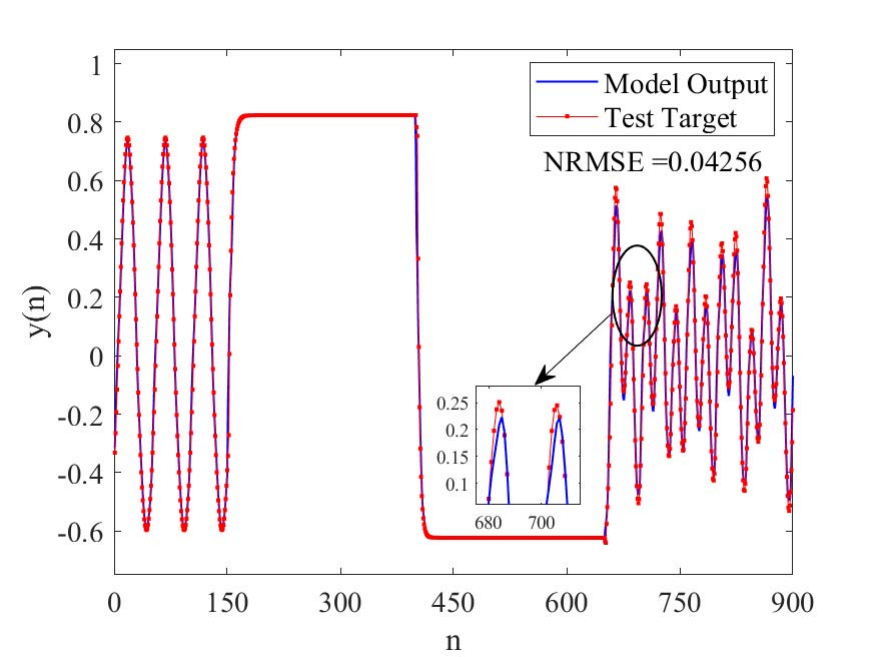}}\\ \vspace{-0.3cm}
    \subfloat[RSCN]{\includegraphics[width=5.5cm]{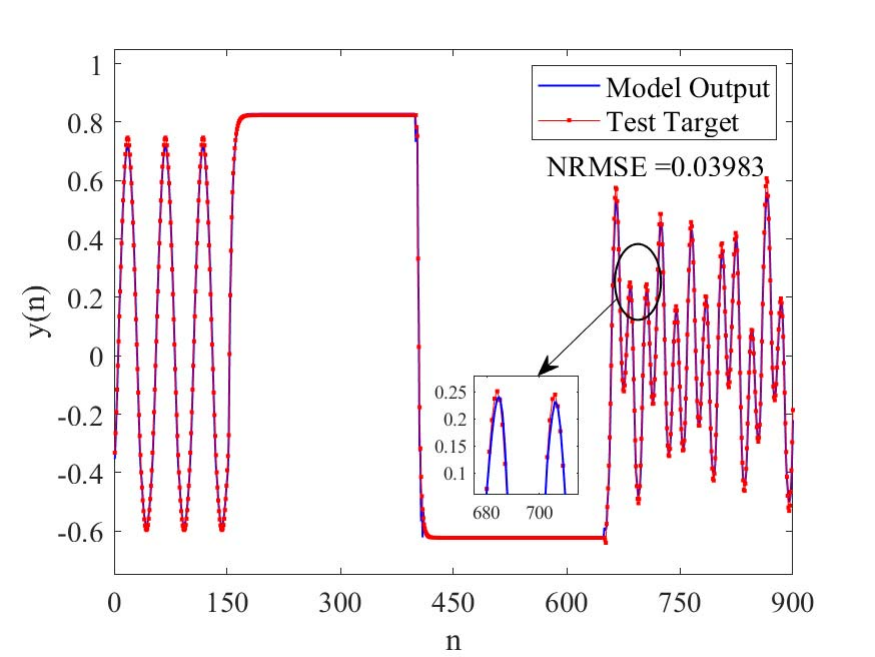}}
	\subfloat[DeepRSCN2]{\includegraphics[width=5.5cm]{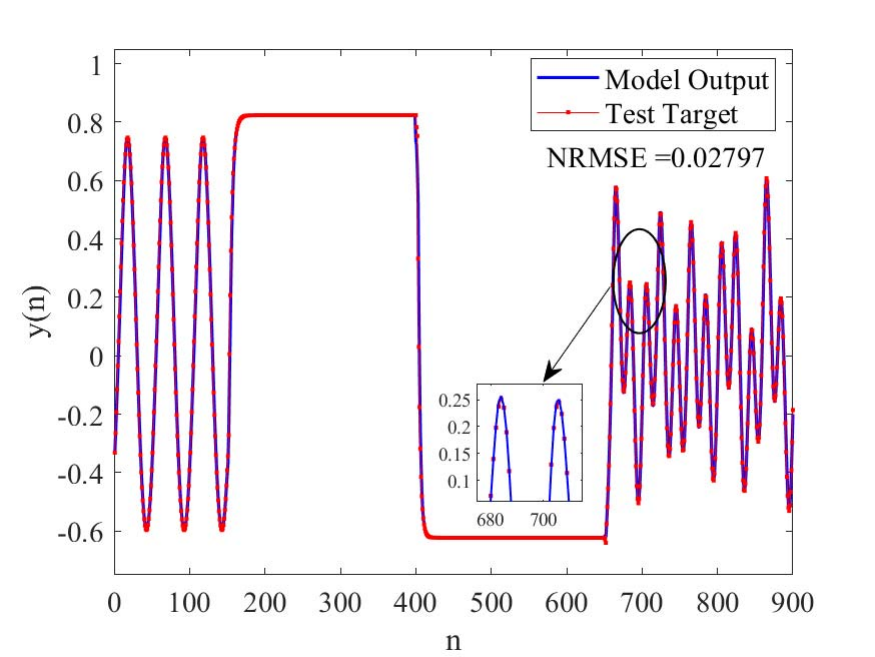}}
    \subfloat[DeepRSCN3]{\includegraphics[width=5.5cm]{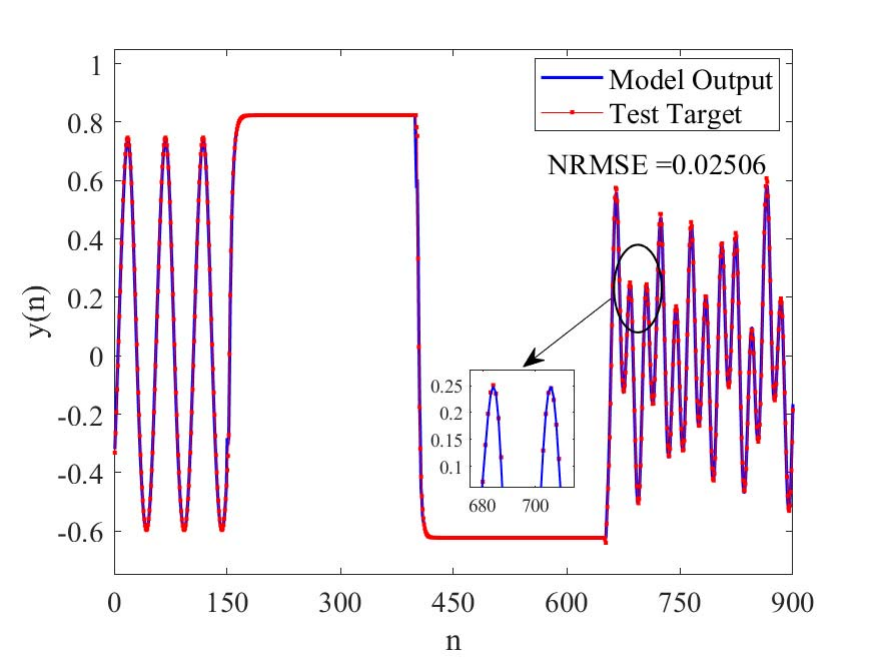}}
	\caption{The prediction curves of each model for the nonlinear system identification task.}
	\label{fig5}
 \vspace{-0.3cm}
\end{figure*}

\begin{figure}[htbp]
\vspace{-0.3cm}
	\centering
        \includegraphics[width=5.5cm]{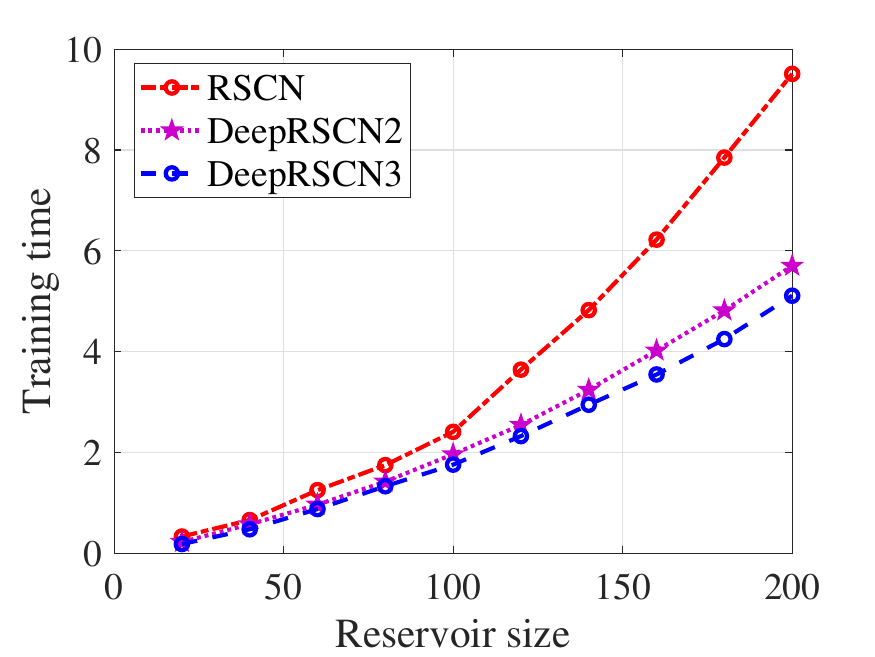}
	\caption{Training time comparison between the RSC-based frameworks with different reservoir sizes on the nonlinear system identification tasks.}
	\label{fig7}
 \vspace{-0.5cm}
\end{figure}

Fig.~\ref{fig511} shows the feature-output correlation distribution across various model architectures. We conducted 30 independent experiments, each utilizing a set of 200 summary nodes, and calculated the average correlation coefficient between the outputs of these nodes and the target outputs. The results clearly indicate that as the number of layers in the model increases, the correlation coefficient significantly strengthens. This phenomenon suggests that deeper models are more effective at capturing the complex patterns and features within the data, which is particularly crucial for nonlinear system identification. By employing DeepRSCNs, we are able to more accurately learn the input-output relationships of the system, thereby enhancing the predictive accuracy.

Fig.~\ref{fig5} displays the prediction curves for each model on the nonlinear system identification task. Obviously, the outputs obtained by DeepRSCN3 exhibit a closer alignment with the desired outputs compared to the other models. Additionally, DeepRSCNs demonstrate a rapid response to dynamic system variations and satisfactory tracking performance, highlighting their significant potential in nonlinear dynamic modelling.

To comprehensively compare and analyze the modelling performance of the proposed methods on the nonlinear system identification task, the simulation results of different models are summarized in Table~\ref{tb2}. It is evident that the RSC-based models outperform the ESN and its deep variants in terms of reservoir size, training, and testing NRMSE. In addition, we can find that the DeepESN3 achieve the lowest training and testing NRMSE with the smallest reservoir size, along with the shortest training time among the RSC-based frameworks, indicating a trend that deeper models perform better. These results suggest that the proposed DeepRSCNs are well-suited for identifying nonlinear systems, especially those with unknown dynamic orders.
\begin{figure}[ht]
\vspace{-0.3cm}
	\begin{center}
		\includegraphics[width=6cm]{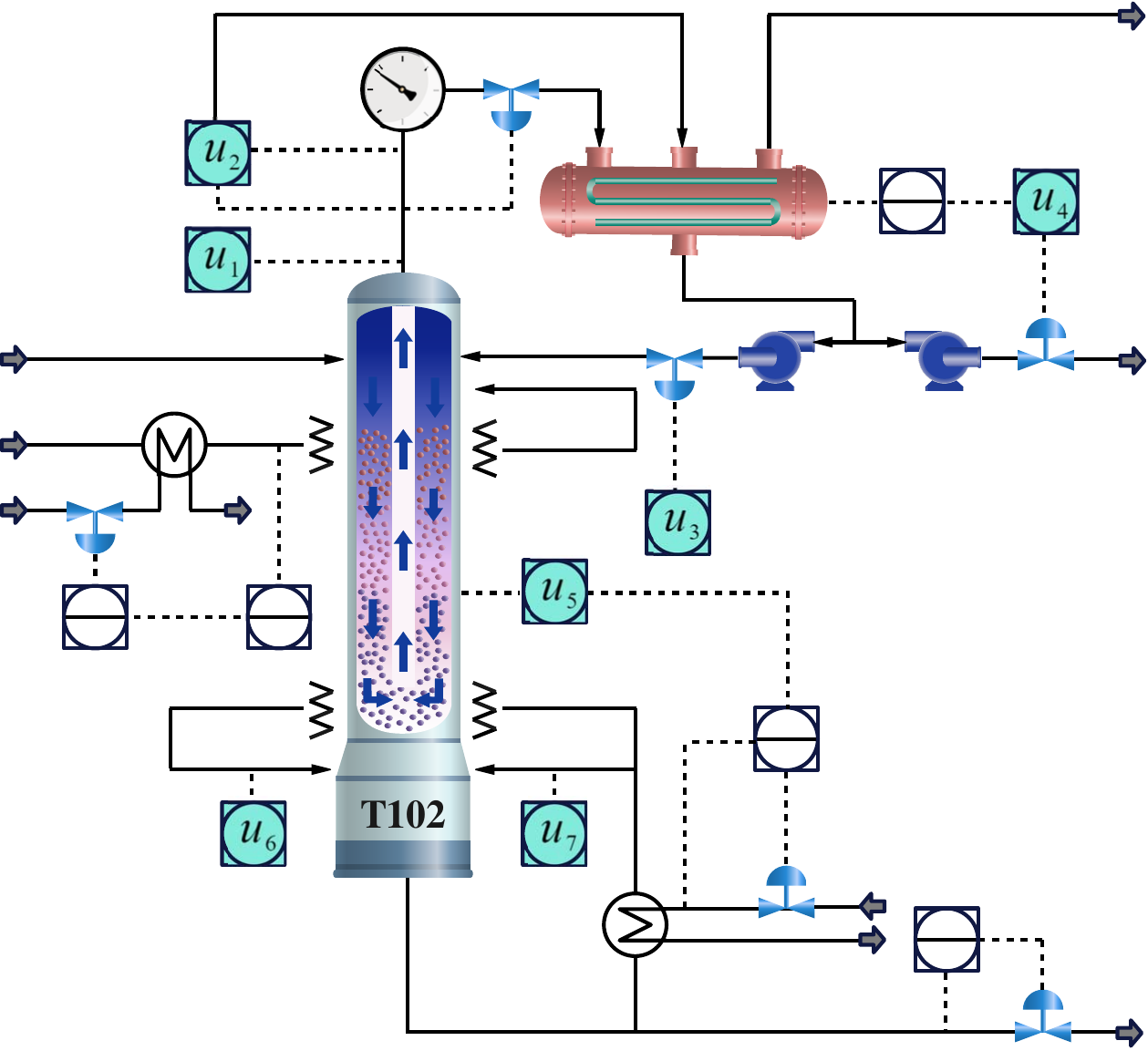}
		\caption{Flowchart of debutanizer column process.}
		\label{fig8}
	\end{center}
\vspace{-0.5cm}
\end{figure}

The training time comparison between the RSC-based frameworks for various reservoir sizes on the two nonlinear system identification tasks is illustrated in Fig.~\ref{fig7}. DeepRSCN3 can achieve excellent training speed for large reservoirs in both tasks, highlighting the superior computational efficiency of multi-layer RSC-based models.
 \vspace{-0.3cm}
 \begin{figure*}[htbp] 
	\centering
	\subfloat[ESN]{\includegraphics[width=8cm]{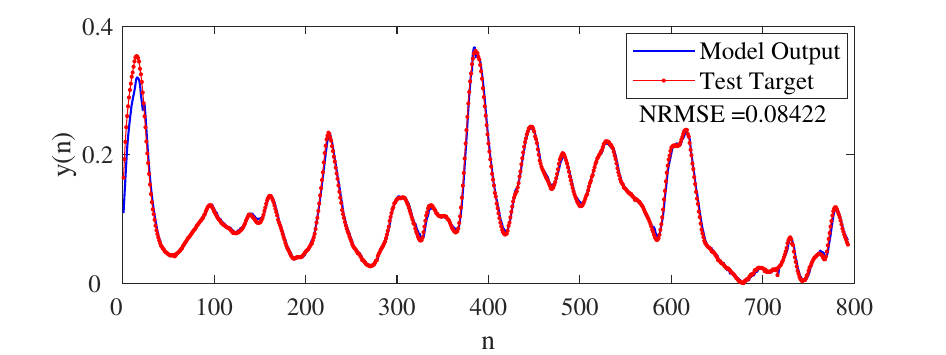}}
	\subfloat[DeepESN2]{\includegraphics[width=8cm]{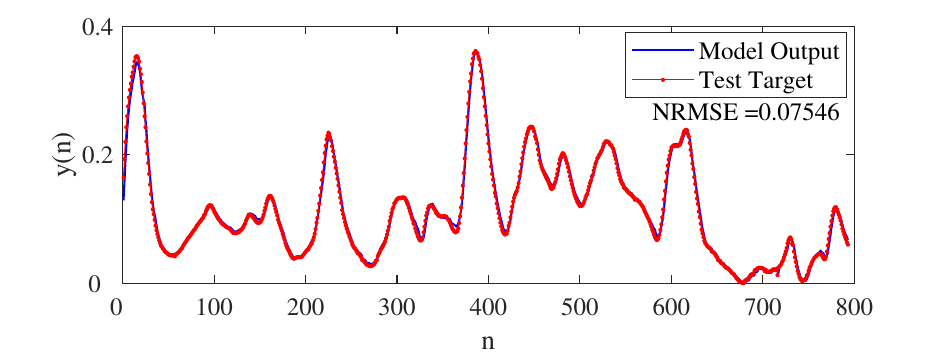}}\\
    \subfloat[DeepESN3]{\includegraphics[width=8cm]{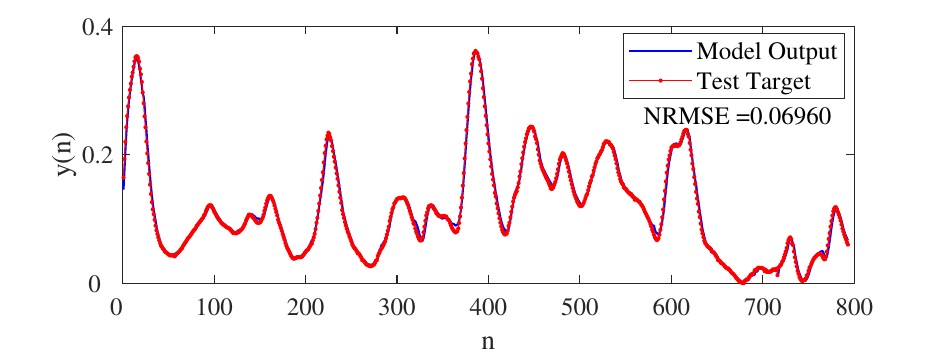}}
	\subfloat[RSCN]{\includegraphics[width=8cm]{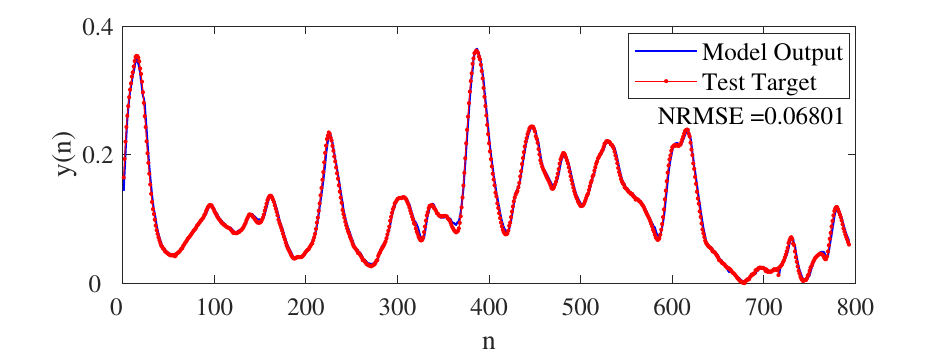}}\\
     \subfloat[DeepRSCN2]{\includegraphics[width=8cm]{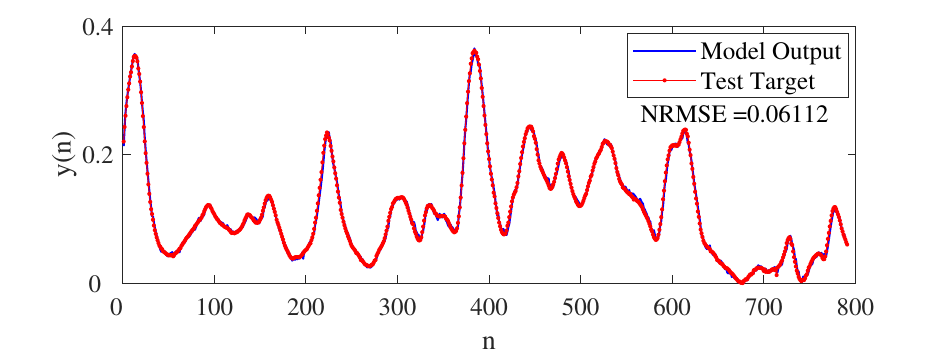}}
	\subfloat[DeepRSCN3]{\includegraphics[width=8cm]{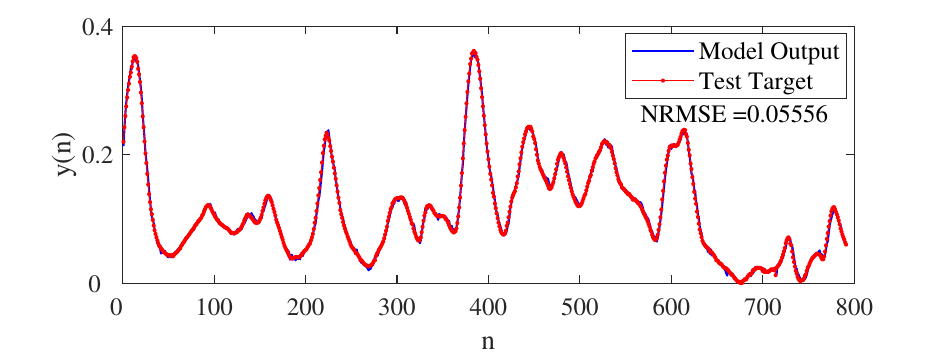}}
	\caption{The prediction curves of each model for debutanizer column process.}
	\label{fig10}
 \vspace{-0.5cm}
\end{figure*}
\subsection{Soft sensing of the butane concentration in the debutanizer column process}
The debutanizer column is a frequently utilized equipment in the oil refining process, designed to separate butane from mixed gas. However, fluctuations in the external environment and alterations in the composition of raw materials may result in discrepancies in direct measurements, which can affect the control and enhancement of the production process. Therefore, it is essential to develop an accurate soft sensing model. As shown in Fig.~\ref{fig8}, several sensors are installed to monitor the process and auxiliary variables generated from these sensors include tower top temperature ${{u}_{1}}$, tower top pressure ${{u}_{2}}$, tower top reflux flow ${{u}_{3}}$, tower top product outflow ${{u}_{4}}$, 6-th tray temperature ${{u}_{5}}$, tower bottom temperature ${{u}_{6}}$, tower bottom pressure ${{u}_{7}}$. The main task is to obtain the butane concentration at the bottom of the tower. In \cite{ref24}, a set of well-designed variables is used to calculate the butane concentration $y$, which can be expressed by \vspace{-0.35cm}

\begin{small}
    \begin{equation} \label{eq37}
\begin{array}{l}
y\left( n \right) = f\left( {{u_1}\left( n \right),} \right.{u_2}\left( n \right),{u_3}\left( n \right),{u_4}\left( n \right),{u_5}\left( n \right),{u_5}\left( {n - 1} \right),\\
{\kern 1pt} {\kern 1pt} {\kern 1pt} {\kern 1pt} {\kern 1pt} {\kern 1pt} {\kern 1pt} {\kern 1pt} {\kern 1pt} {\kern 1pt} {\kern 1pt} {\kern 1pt} {\kern 1pt} {\kern 1pt} {\kern 1pt} {\kern 1pt} {\kern 1pt} {\kern 1pt} {\kern 1pt} {\kern 1pt} {\kern 1pt} {\kern 1pt} {\kern 1pt} {\kern 1pt} {\kern 1pt} {\kern 1pt} {\kern 1pt} {\kern 1pt} {\kern 1pt} {\kern 1pt} {\kern 1pt} {\kern 1pt} {\kern 1pt} {\kern 1pt} {\kern 1pt} {\kern 1pt} {\kern 1pt} {\kern 1pt} {\kern 1pt} {\kern 1pt} {\kern 1pt} {\kern 1pt} {\kern 1pt} {\kern 1pt} {\kern 1pt} {\kern 1pt} {\kern 1pt} {\kern 1pt} {\kern 1pt} {\kern 1pt} {u_5}\left( {n - 2} \right),{u_5}\left( {n - 3} \right),\left( {{u_1}\left( n \right) + {u_2}\left( n \right)} \right)/2,\\
{\kern 1pt} {\kern 1pt} {\kern 1pt} {\kern 1pt} {\kern 1pt} {\kern 1pt} {\kern 1pt} {\kern 1pt} {\kern 1pt} {\kern 1pt} {\kern 1pt} {\kern 1pt} {\kern 1pt} {\kern 1pt} {\kern 1pt} {\kern 1pt} {\kern 1pt} {\kern 1pt} {\kern 1pt} {\kern 1pt} {\kern 1pt} {\kern 1pt} {\kern 1pt} {\kern 1pt} {\kern 1pt} {\kern 1pt} {\kern 1pt} {\kern 1pt} {\kern 1pt} {\kern 1pt} {\kern 1pt} {\kern 1pt} {\kern 1pt} {\kern 1pt} {\kern 1pt} {\kern 1pt} {\kern 1pt} {\kern 1pt} {\kern 1pt} {\kern 1pt} {\kern 1pt} {\kern 1pt} {\kern 1pt} {\kern 1pt} {\kern 1pt} {\kern 1pt} {\kern 1pt} \left. {{\kern 1pt} {\kern 1pt} {\kern 1pt} y\left( {n - 1} \right),y\left( {n - 2} \right),y\left( {n - 3} \right),y\left( {n - 4} \right)} \right).
\end{array}
\end{equation}
\end{small}Considering order uncertainty, the output is predicted by  $\left[ {{u}_{1}}\left( n \right), \right.$$\left. {{u}_{2}}\left( n \right),{{u}_{3}}\left( n \right),{{u}_{4}}\left( n \right),{{u}_{5}}\left( n \right),y\left( n-1 \right) \right]$ in this simulation. A total of 2394 samples are generated from real-time sampling process, and the first 1500 samples are used for training, the following 894 samples are used for testing, and the Gaussian white noise is added to the testing set to generate the validation set. The first 100 samples of each set are washed out.
\begin{figure}[htbp]
\vspace{-0.3cm}
	\begin{center}
		\includegraphics[width=7.5cm]{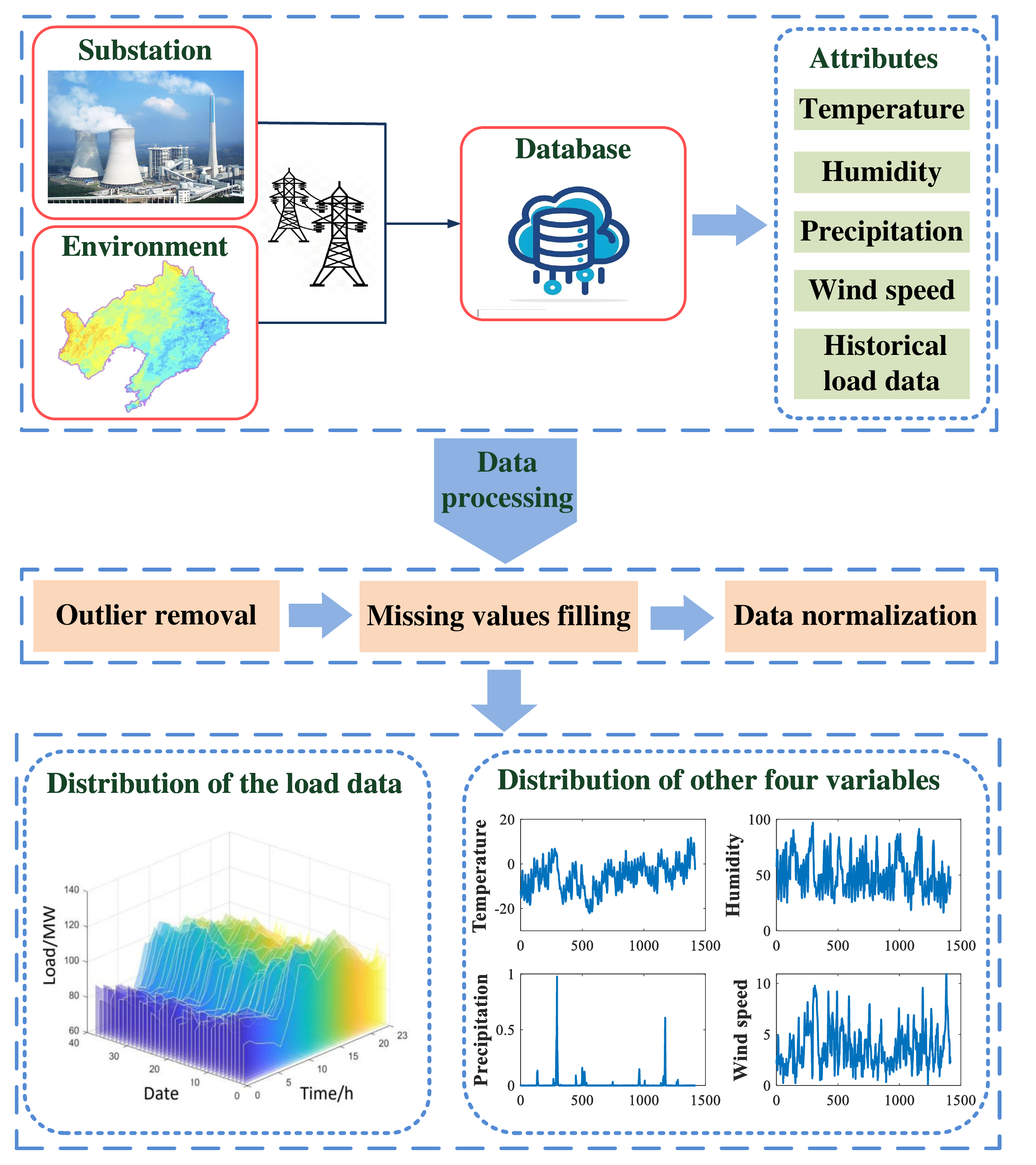}
		\caption{Flowchart of the data collection and processing for the short-term power load forecasting.}
		\label{fig9}
	\end{center}
 \vspace{-0.4cm}
\end{figure}

Fig.~\ref{fig10} illustrates the prediction curves of each model for the dehumanizer column process. The prediction accuracy of the RSCN framework is significantly higher than that of the ESN. Furthermore, as the model depth increases, the fitting performance also improves. These results suggest that a deeper structure can effectively enhance the model's ability to approximate nonlinear relationships, making it particularly suitable for complex industrial processes.

\subsection{Short-term power load forecasting}
\begin{figure*} 
	\centering
	\subfloat[ESN]{\includegraphics[width=8cm]{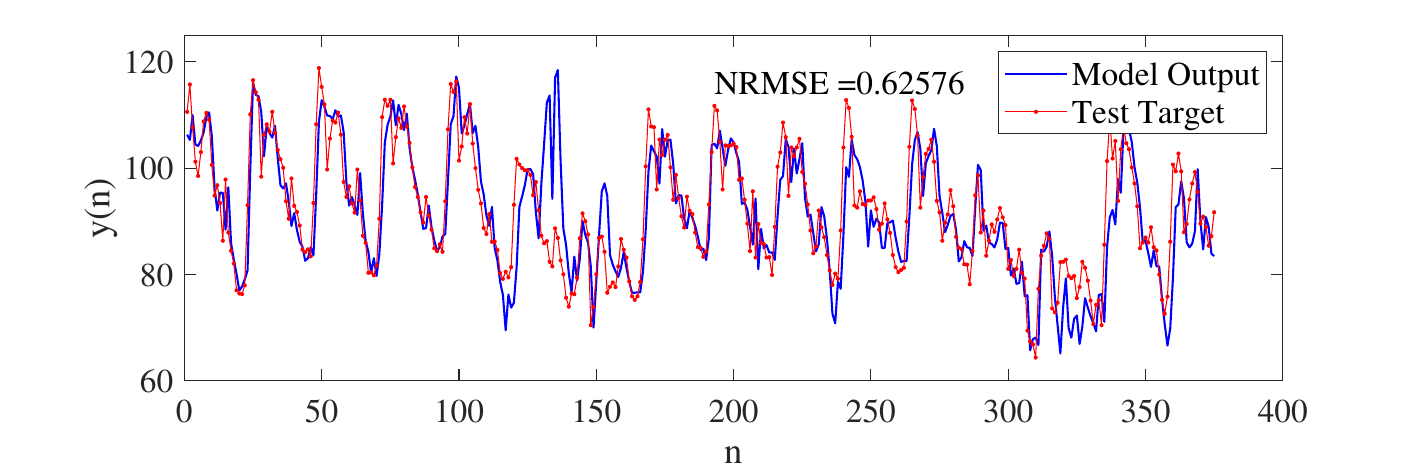}}
	\subfloat[DeepESN2]{\includegraphics[width=8cm]{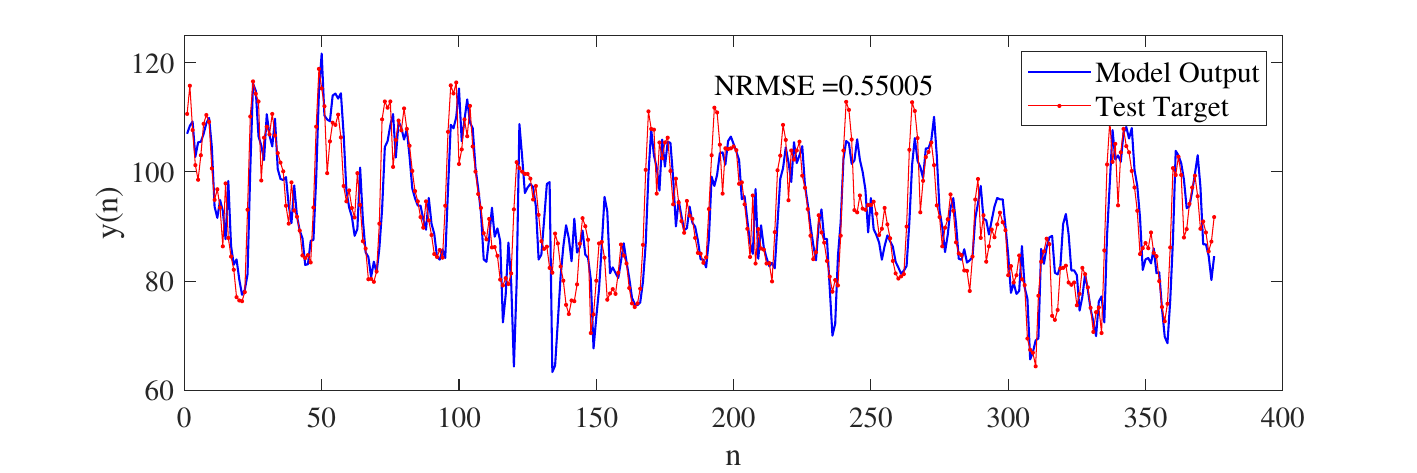}}\\
    \subfloat[DeepESN3]{\includegraphics[width=8cm]{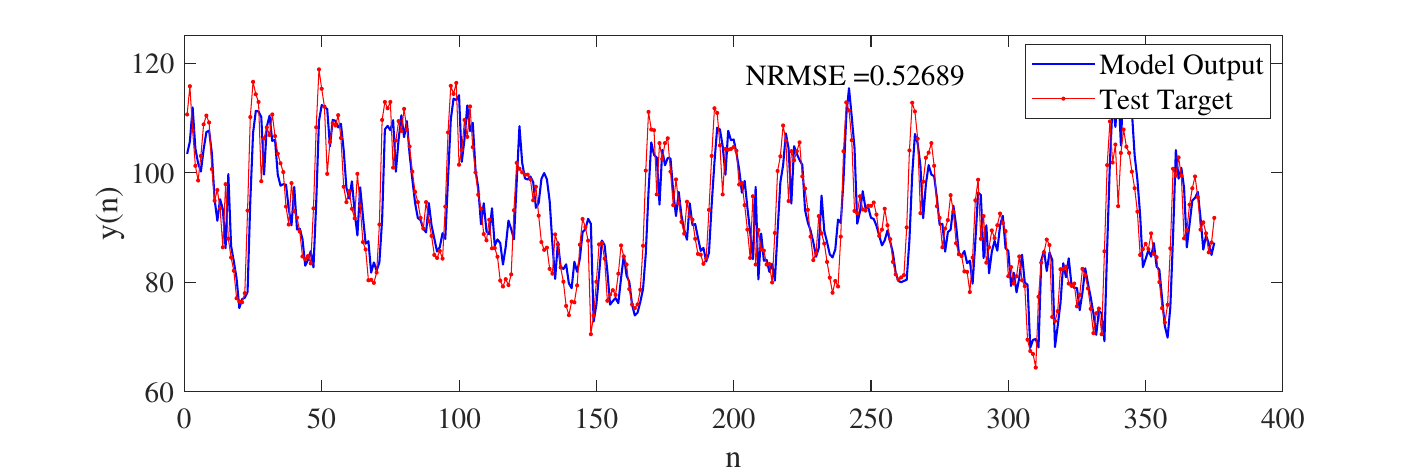}}
	\subfloat[RSCN]{\includegraphics[width=8cm]{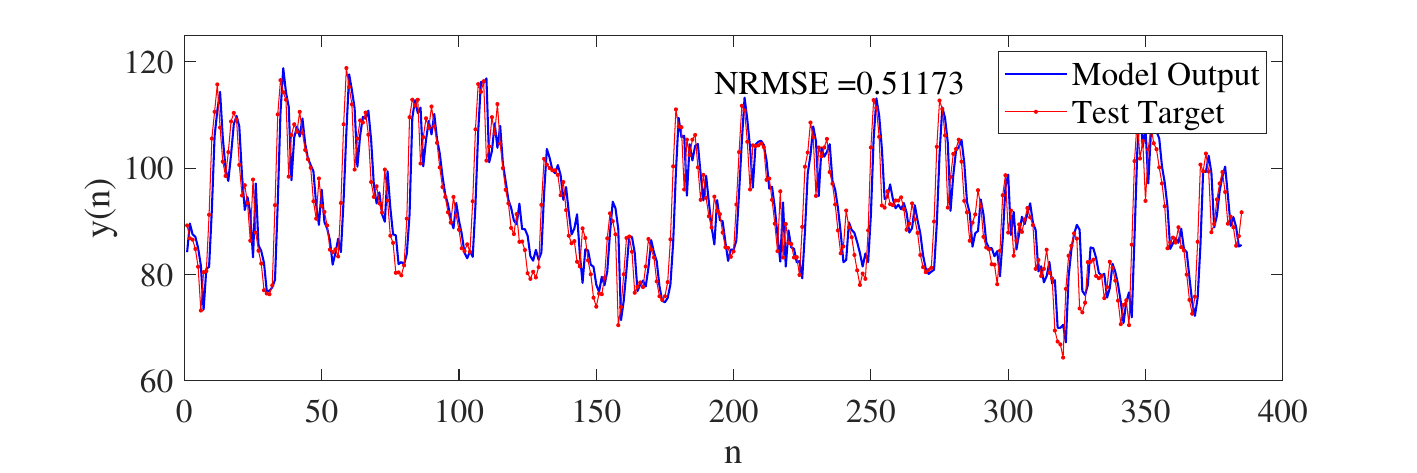}}\\
     \subfloat[DeepRSCN2]{\includegraphics[width=8cm]{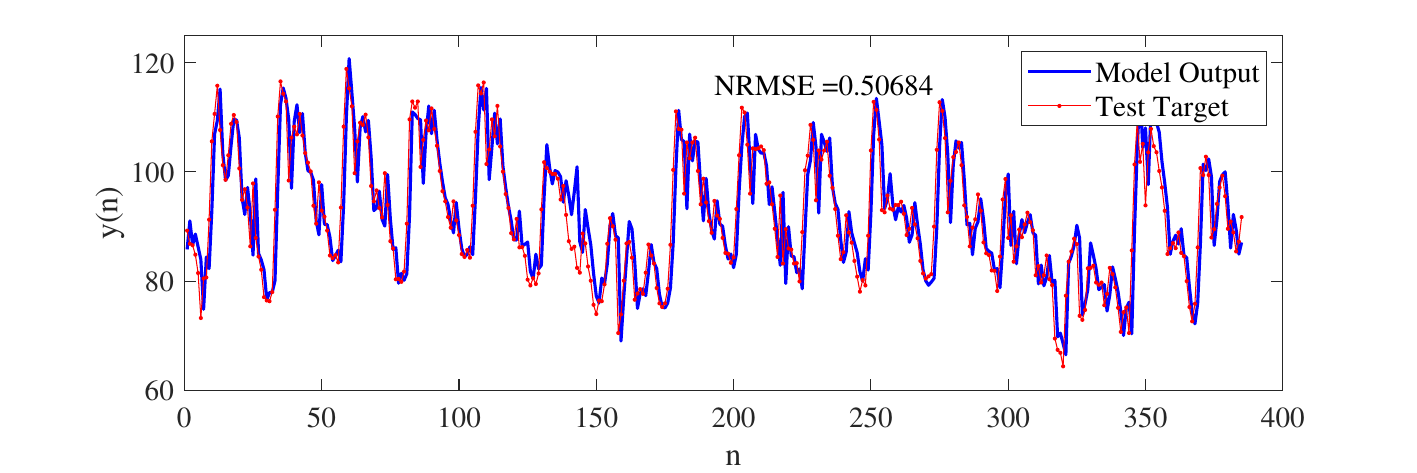}}
	\subfloat[DeepRSCN3]{\includegraphics[width=8cm]{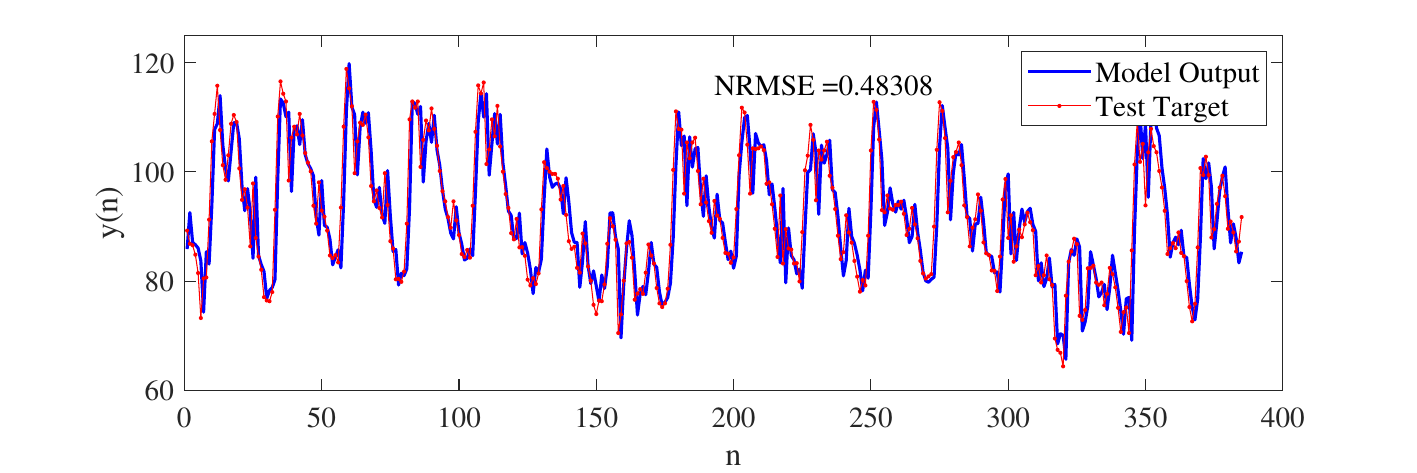}}
	\caption{The prediction curves of each model for the short-term power load.}
	\label{fig11}
 \vspace{-0.5cm}
\end{figure*}
\begin{figure*}[htbp]
\vspace{-0.3cm}
	\centering
	\subfloat[Case 1]{\includegraphics[width=7cm]{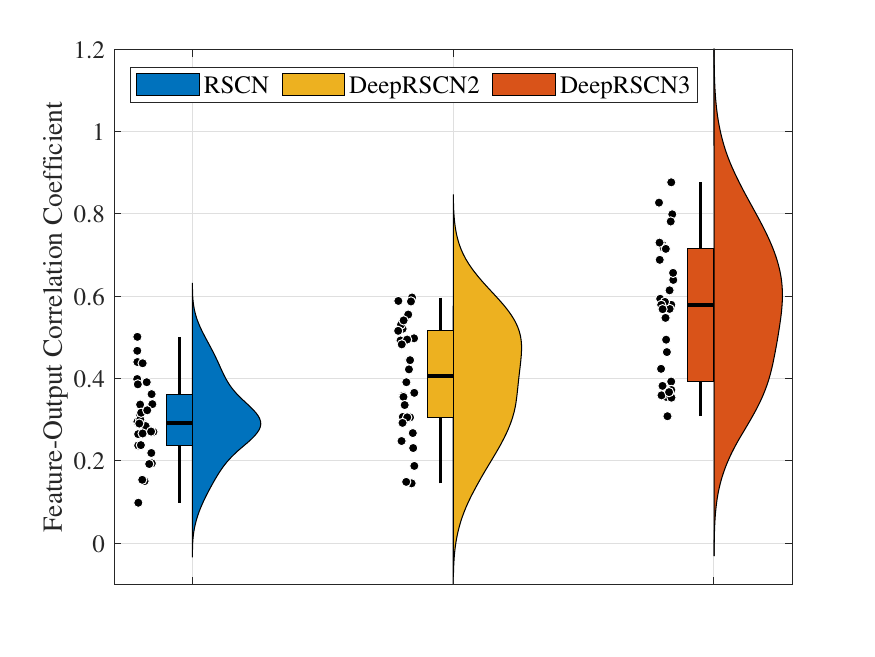}}
	\subfloat[Case 2]{\includegraphics[width=7cm]{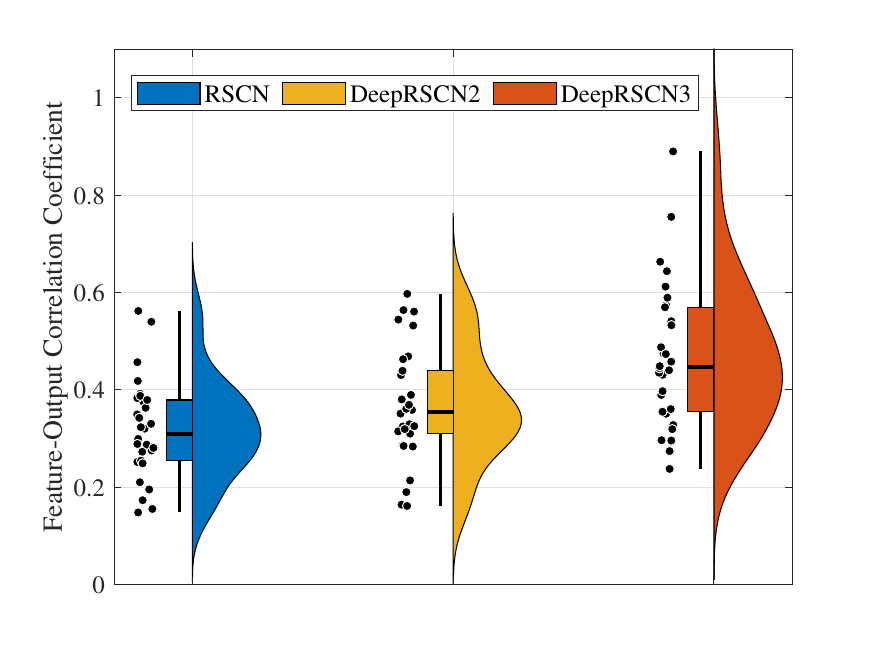}}
		\caption{The correlation between reservoir outputs and target outputs across
varying model depths on the two industry cases.}
		\label{fig111}
   \vspace{-0.35cm}
\end{figure*}
Short-term power load forecasting plays a critical role in ensuring the reliable and efficient operation of power systems, as well as in cost reduction. This study utilizes load data from a 500kV substation in Liaoning Province, China, collected hourly from January to February 2023 across 59 days. Environmental factors such as temperature ${u_1}$, humidity ${u_2}$, precipitation ${u_3}$, and wind speed ${u_4}$ were considered to predict power load $y$. The process of data collection and analysis for short-term electricity load forecasting is depicted in Fig.~\ref{fig9}. The dataset comprises 1415 samples, with 1000 samples allocated for training and 415 for testing. Gaussian noise is added to the testing set to generate the validation set. Taking into account the order uncertainty, $\left[ {{u}_{1}}\left( n \right), \right.$$\left. {{u}_{2}}\left( n \right),{{u}_{3}}\left( n \right),{{u}_{4}}\left( n \right),y\left( n-1 \right) \right]$ is used to predict $y\left( n \right)$ in the experiment. The first 30 samples of each set are washed out.

Fig.~\ref{fig11} depicts the prediction curves of each model for the short-term power load forecasting. It is obvious that the outputs generated by DeepRSCNs have a higher fitting degree with the desired outputs in comparison to the other models, thereby verifying the feasibility of the proposed approach for predicting industrial process parameters.

Fig.~\ref{fig111} exhibits the feature-output correlation distribution across various model architectures in two industrial case studies. The experimental results clearly indicate that deeper network models demonstrate a stronger correlation between the reservoir outputs and the target outputs. This enhanced correlation suggests that deeper architectures are more effective in capturing intricate relationships within the data. By improving feature representation capabilities, DeepRSCNs can maintain high prediction accuracy under varying operational conditions, learn key dynamic features, and respond promptly to changes in system states. This ability contributes to increased stability in process control and enhances the optimization efficiency of industrial processes.

\begin{figure*}[htbp]
\vspace{-0.5cm}
	\centering
	\subfloat[Case 1]{\includegraphics[width=5.5cm]{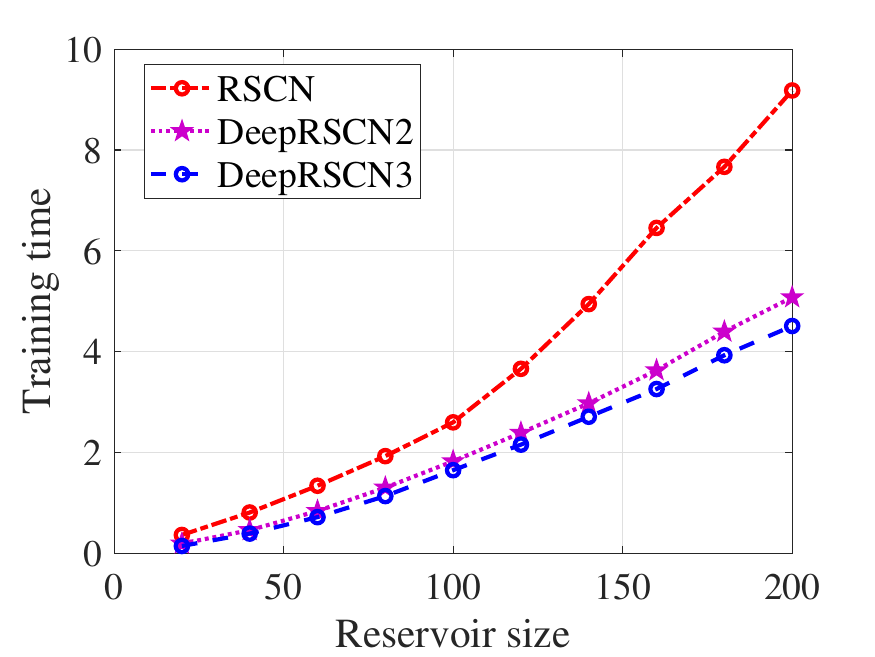}}
	\subfloat[Case 2]{\includegraphics[width=5.5cm]{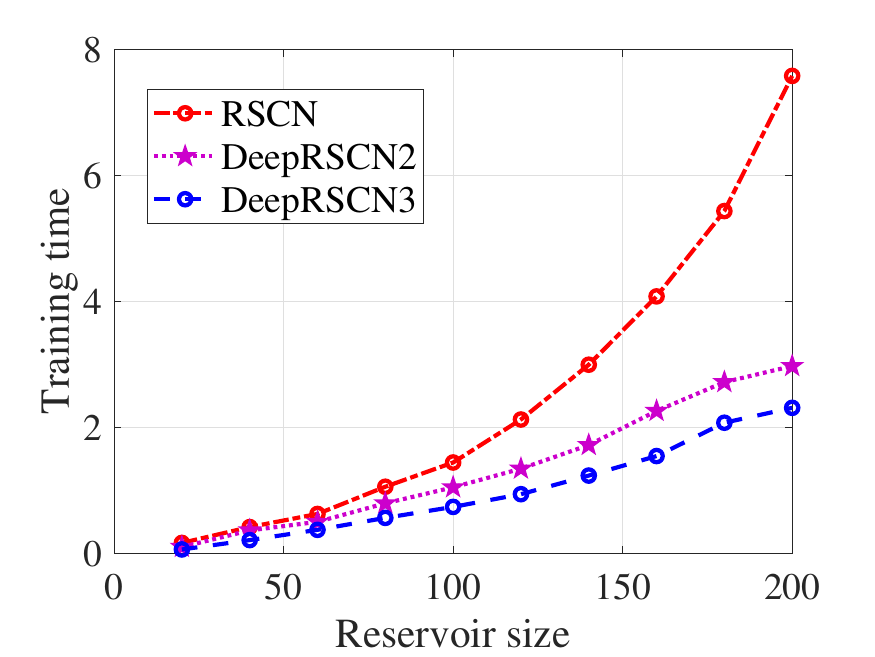}}
		\caption{Training time comparison between the RSC-based frameworks with different reservoir sizes on the two industry cases.}
		\label{fig12}
   \vspace{-0.35cm}
\end{figure*}

To facilitate the comparison of the modelling performance of different models, the simulation results are summarized in Table~\ref{tb3}. It can be seen that DeepRSCNs show smaller NRMSE in both training and testing performance, and their reservoir topologies are more compact. Fig.~\ref{fig12} gives the training time consumption of the three RSC-based frameworks for various reservoir sizes, where DeepRSCNs exhibit superior computational efficiency. These results demonstrate that DeepRSCNs can reduce computational costs while enhancing the model performance, which have great potential for tackling complex dynamic modelling problems in practical industrial processes. \vspace{-0.2cm}

\begin{table*}[]
\caption{Performance comparison of different models on the two industry cases.} \label{tb3}
\centering
\begin{tabular}{cccccc}
\hline
Datasets                & Models    & Reservoir size & Training time (s)           & Training NRMSE           & Testing NRMSE            \\ \hline
\multirow{6}{*}{Case 1} & ESN       & 213            & 0.40819±0.12035          & 0.04085±0.00137          & 0.08427±0.01144          \\
                        & DeepESN2  & 100-89         & 0.37296±0.05362          & 0.03060±0.00106          & 0.07554±0.00378          \\
                        & DeepESN3  & 60-60-37       & \textbf{0.35203±0.07601} & 0.03024±0.00094          & 0.06970±0.00855          \\
                        & RSCN      & 87             & 2.23734±0.20122          & \textbf{0.02734±0.00100} & 0.06793±0.00411          \\
                        & DeepRSCN2 & 50-22          & 1.04856±0.14261          & 0.02906±0.00117          & 0.06095±0.00258          \\
                        & DeepRSCN3 & 30-30-5        & 0.76352±0.10127          & 0.02801±0.00108          & \textbf{0.05552±0.00299} \\ \hline
\multirow{6}{*}{Case 2} & ESN       & 103            & 0.11881±0.01548          & 0.24338±0.02189          & 0.64522±0.07172          \\
                        & DeepESN2  & 80-32          & 0.11292±0.02028          & 0.23761±0.01938          & 0.60285±0.07384          \\
                        & DeepESN3  & 40-40-11       & \textbf{0.09219±0.00783} & 0.21928±0.01827         & 0.59285±0.04633          \\
                        & RSCN      & 65             & 0.68257±0.03141          & 0.19254±0.02047          & 0.51078±0.03685          \\
                        & DeepRSCN2 & 40-18          & 0.48392±0.03302          & 0.19098±0.02375          & 0.50329±0.01769          \\
                        & DeepRSCN3 & 20-20-13       & 0.36964±0.02938          & \textbf{0.19037±0.02019} & \textbf{0.48997±0.02983} \\ \hline
\end{tabular}
\end{table*}
\subsection{Discussions}
DeepRSCNs inherit the strengths of the original RSCNs, such as data-dependent parameter selection,  theoretical basis for structural design, and strong nonlinear approximation capability. The incorporation of deep learning techniques further improves the model performance by enhancing data representation, enabling more efficient capture and extraction of high-level features. Consequently, DeepRSCNs demonstrate heightened flexibility and expressive power, allowing them to effectively capture complex patterns in the process data. Experimental results show that the proposed DeepRSCNs can also improve the model compactness and learning efficiency. These findings verify the superior adaptability of DeepRSCNs to diverse complex data distributions and features, highlighting their great potential in modelling nonlinear dynamic systems.\vspace{-0.2cm}

\section{Conclusion}
This paper presents a deep version of RSCN with the universal approximation property for modelling nonlinear dynamic systems. The effectiveness of the proposed approach is validated across four nonlinear dynamic modelling tasks. Experimental results demonstrate that DeepRSCNs can achieve sound performance in terms of modelling efficiency, learning capability, and prediction accuracy. By analyzing the performance of RSC-based frameworks with varying numbers of layers, we found that deep models consistently outperform single-layer models. Such a finding clearly indicates that the deep structure can enhance the model's feature representation capabilities through multi-level feature extraction and representation, thereby more effectively capturing complex patterns and nonlinear relationships in the data, leading to more accurate and reliable prediction results.

There are multiple opportunities for further developing the proposed framework, both in theoretical aspects and practical applications. For example, researchers could investigate different types of supervisory mechanisms or alternative learning schemes for random representation. On the other hand, future studies could explore the self-organizing version of DeepRSCNs to enhance the model's online self-learning capability in dealing with unknown non-stationary data streams.

\end{CJK}


\begin{thebibliography}{99}
\bibitem{ref1}Y. LeCun, Y. Bengio, and G. Hinton, “Deep learning," Nature, vol. 521, no. 7553, pp. 436–444, May 2015.

\bibitem{ref2}J. Schmidhuber, “Deep learning in neural networks: an overview," Neural Networks, vol. 61, pp. 85–117, Jan. 2015.

\bibitem{ref201}D. Wang and K. Liu, “An integrated deep learning-based data fusion and degradation modeling method for improving prognostics," IEEE Trans. Autom. Sci. Eng., vol. 21, no. 2, pp. 1713-1726, Apr. 2024.

\bibitem{ref202}S. Gao, H. Zhang, Z. Wang, H. Huang and H. Yan, “Data-driven jnjection attack against discrete-time intelligent automation systems with slowly time-varying delays," IEEE Trans. Autom. Sci. Eng., Early access (online available), Oct. 2023.

\bibitem{ref203}J. Liu, X. Yang, H. Zhang, Z. Wang and H. Yan, “A time-delay modeling approach for data-driven predictive control of continuous-time systems," IEEE Trans. Autom. Sci. Eng., Early access (online available), Aug. 2024.

\bibitem{ref3}R. Pascanu, C. Gulcehre, K. Cho, and Y. Bengio, “How to construct deep recurrent neural networks," arXiv:1312.6026v5, 2014.

\bibitem{ref4}M. Hermans and B. Schrauwen, “Training and analysing deep recurrent neural networks," in Advances in Neural Information Processing Systems, vol. 26, pp. 190–198, 2013.

\bibitem{ref6}M. Lukosevicius and H. Jaeger, “Reservoir computing approaches to recurrent neural network training," Comput. Sci. Rev., vol. 3, no. 3, pp. 127–149, Aug. 2009.

\bibitem{ref7}S. Scardapane and D. Wang, “Randomness in neural networks: an overview," Wiley Interdiscip. Rev.: Data Min. Knowl. Discovery, vol. 7, no. e1200, Feb. 2017.

\bibitem{ref711}D. Wang, “Editorial: randomized algorithms for training neural networks," Inf. Sci., vol. 364–365, pp. 126–128, Oct. 2016.

\bibitem{ref712}J. Huang, Y. Cao, C. Xiong and H.-T. Zhang, “An echo state gaussian process-based nonlinear model predictive control for pneumatic muscle actuators," IEEE Trans. Autom. Sci. Eng., vol. 16, no. 3, pp. 1071-1084, Jul. 2019.

\bibitem{ref8}H. Jaeger, “The echo state approach to analysing and training recurrent neural networks-with an erratum note,” German Nat. Res. Center Inf. Technol., Bonn, Germany, Tech. Rep. GMD, 148, 2001.

\bibitem{ref9}W. Maass, T. Natschläger, and H. Markram, “Real-time computing without stable states: a new framework for neural computation based on perturbations,” Neural Comput., vol. 14, no. 11, pp. 2531-2560, Nov. 2002.

\bibitem{ref001}R. Eldan and O. Shamir, “The power of depth for feedforward neural networks,” in Conference on Learning Theory, PMLR, pp. 907-940, 2016.

\bibitem{ref10}Z. K. Malik, A. Hussain, and Q. Wu, “Multilayered echo state machine: a novel architecture and algorithm,” IEEE Trans. Cybern., vol. 47, no. 4, pp. 946-959, Apr. 2017.

\bibitem{ref11}C. Gallicchio, A. Micheli, and L. Pedrelli, “Deep reservoir computing: a critical experimental analysis,” Neurocomputing, vol. 268, pp. 87–99, Dec. 2017.

\bibitem{ref12}T. Akiyama and G. Tanaka, “Computational efficiency of multi-step learning echo state networks for nonlinear time series prediction,” IEEE Access, vol. 10, pp. 28535-28544, Mar. 2022.

\bibitem{ref13}C. Gallicchio and A. Micheli, “Echo state property of deep reservoir computing networks,” Cognit. Comput., vol. 9, no. 3, pp. 337–350, May 2017.

\bibitem{ref14}C. Gallicchio and A. Micheli, “Deep echo state network: a brief survey,” arxiv:1712.04323v4, Sep. 2020.

\bibitem{ref16}D. Wang and M. Li, “Deep stochastic configuration networks with universal approximation property,” in 2018 International Joint Conference on Neural Networks (IJCNN), Rio de Janeiro, Brazil, pp. 1-8, 2018.

\bibitem{ref17}M. Li and D. Wang, “Insights into randomized algorithms for neural networks: practical issues and common pitfalls,” Inf. Sci., vol. 382–383, pp. 170–178, Mar. 2017.

\bibitem{ref18}D. Wang and M. Li, “Stochastic configuration networks: fundamentals and algorithms,” IEEE Trans. Cybern., vol. 47, no. 10, pp. 3466-3479, Oct. 2017.

\bibitem{ref181}D. Wang and G. Dang, “Recurrent stochastic configuration networks for temporal data analytics,” arXiv: 2406.16959v2, Sep. 2024.

\bibitem{ref19}D. Wang and M. Li, “Robust stochastic configuration networks with kernel density estimation for uncertain data regression,” Inf. Sci., vol. 412–413, pp. 210–222, Oct. 2017.

\bibitem{ref20}K. Li, J. Qiao, and D. Wang, “Online self-learning stochastic configuration networks for nonstationary data stream analysis,” IEEE Trans. Ind. Inf., vol. 20, no. 3, pp. 3222-3231, Mar. 2024.

\bibitem{ref21}H. Jaeger, M. Lukoševičius, D. Popovici, and U. Siewert, “Optimization and applications of echo state networks with leaky-integrator neurons,” Neural Networks, vol. 20, no. 3, pp. 35-352, Apr. 2007.

\bibitem{ref22}C. H. Valencia, M. Vellasco, and K. Figueiredo, “Echo state networks: novel reservoir selection and hyperparameter optimization model for time series forecasting,” Neurocomputing, vol. 545, no. 126317, Aug. 2023.

\bibitem{ref23}G. Goodwin and K. Sin, “Adaptive filtering prediction and control,” Courier Corporation, 2014.

\bibitem{ref24}L. Fortuna, S. Graziani, and M. G. Xibilia, “Soft sensors for product quality monitoring in debutanizer distillation columns,” Control Eng. Pract., vol. 13, no. 4, pp. 499-508, Apr. 2005.
\end{thebibliography}
\end{document}